\documentclass{article}

\usepackage[margin=1.25in]{geometry}
\usepackage{enumitem}
\usepackage{color, colortbl}
\usepackage[utf8]{inputenc} 
\usepackage[T1]{fontenc}    
\usepackage{hyperref}       
\usepackage{url}            
\usepackage{booktabs}       
\usepackage{amsfonts}       
\usepackage{nicefrac}       
\usepackage{microtype}      
\usepackage{caption}

\usepackage{fp}

\usepackage{float}
\usepackage{tikz}
\usetikzlibrary{calc}
\usetikzlibrary{shapes.geometric}
\usetikzlibrary{dsp}
\usetikzlibrary{chains}

\usepackage{xifthen}
\usepackage{amsmath}
\usepackage{amsthm}
\usepackage{amssymb}

\usepackage{cite}

\usetikzlibrary{positioning}

\newcommand{\realnumber}{\mathbb{R}}

\newcommand{\enc}{e}

\newtheorem{theorem}{Theorem}
\newtheorem{lemma}{Lemma}

\newcommand\cisymbol{\perp\!\!\!\perp}

\newcommand{\ci}[3]{(#1 \cisymbol #2 | #3)}

\title{Lattice Representation Learning}

\author{%
  Luis A. Lastras\\
  IBM Research AI\\
  T.J. Watson Research Center\\
  Yorktown Heights, NY, 10598 \\
  \texttt{lastrasl@us.ibm.com} \\
}

\begin{document}

\maketitle

\begin{abstract}
In this article we introduce theory and algorithms for learning discrete representations that take on a lattice that is embedded in an Euclidean space. Lattice representations possess an interesting combination of properties: a) they can be computed explicitly using lattice quantization, yet they can be learned efficiently using the ideas we introduce in this paper, b) they are highly related to Gaussian Variational Autoencoders, allowing designers familiar with the latter to easily produce discrete representations from their models and c) since lattices satisfy the axioms of a group, their adoption can lead into a way of learning simple algebras for modeling binary operations between objects through symbolic formalisms, yet learn these structures also formally using differentiation techniques. This article will focus on laying the groundwork for exploring and exploiting the first two properties, including a new mathematical result linking expressions used during training and inference time and experimental validation on two popular datasets.

\end{abstract}

\section{Introduction}

    
A statistician has observations $x_1,x_2,\cdots,x_n$ from an alphabet $\mathcal{X}$ and wishes to obtain a model $p_X(x)$ for these, as well as other unseen potential observations. The statistician believes that it is reasonable to think of $p_X(x)$ as a marginal of a distribution $p_{X,Z}(x,z) = p_Z(z) p_{X|Z}(x|z)$:
\begin{eqnarray}
p_X(x) = E_{Z \sim p_Z} \left[ p_{X|Z}(x | Z) \right].
\label{eq:lvm}
\end{eqnarray}
Furthermore, the statistician has reasons to believe that $z \in \mathcal{Z}$ should be thought of as a discrete variable. Still, the statistician can't afford engage in a complex application specific modeling endeavor, and therefore wishes to use unsupervised methods. The statistician is not only interested in learning a factorized latent variable model as in (\ref{eq:lvm}); she wants to be able to learn a discrete representation for a given object $x \in \mathcal{X}$. Because she believes $\mathcal{Z}$ to be relatively large to estimate (\ref{eq:lvm}) directly, she will use Variational Inference and the Evidence Lower BOund (ELBO) as used in Variational Auto Encoders (VAEs) \cite{peterson_anderson:mean_field_1987, parisi:statistical_field_theory, saul1996exploiting, saul1996mean, jaakkola1997variational, ghahramani1997factorial, jordan1999introduction,hinton1993keeping, neal1998view, dempster1977maximum, kingma:VAE} which introduces a ``helper'' conditional distribution $Q_{Z|X}$ and states that
\begin{eqnarray}
\log p_X(x) \geq E_{Z \sim Q_{Z|X}(\cdot | x)} \left[ \log \frac{p_Z(Z)}{Q_{Z|X}(Z|x)} \right] + E_{Z \sim Q_{Z|X}(\cdot | x)} \left[ \log p_{X|Z}(x | Z) \right].
\label{eq:elbo}
\end{eqnarray}



%
In the ELBO, the first term is called the representation cost and the second one is called the reconstruction cost. In its most general form, optimizing the ELBO leads to a \emph{stochastic} representation through $Q_{Z|X}$. If we intend to use the representation as part of, say a system for compressing data from $\mathcal{X}$ or as a component of a symbolic system (e.g. a planner), a stochastic representation is not that useful.  In this article, we adopt the viewpoint  that a discrete representation should be modeled as a digital communication channel between two parts of a computational network \cite{shannon1948mathematical, Sha59, berger1971rate}, where the channel conveys an explicit discrete representation for $x$ whose average representation cost, in bits, matches that of the ELBO's representation cost. As an example, in VQ-VAE \cite{van2017neural} the approximate posterior $Q_{Z|X}$ is deterministic and thus it satisfies this requirement. The discussion on the distinction between stochastic and explicit representations is much deeper than what we are making apparent here and merits a separate discussion; a few remarks in this direction can be found in the Appendix's Subsection \ref{ss:digression}.

The purpose of this article is to introduce a type of discrete representation with close ties to the Gaussian Variational Auto Encoder \cite{kingma:VAE} and VQ-VAE that also meets this requirement, has interesting additional properties that are useful for designing training algorithms and analyzing performance. In these representations, the alphabet $\mathcal{Z}$ is a \emph{lattice} \cite{conway2013sphere}; two such example lattices are illustrated in Figure \ref{fig:lattice}. A lattice is an example of a group, this is, a set together with a binary operation that satisfies the axioms of closure, associativity, identity and invertibility. This property could be used to construct representations in which binary operations between object latent properties are modeled explicitly, leading the way to new ways to learn formal algebras from data. Although in our article we do not directly exploit this observation, this is a key motivation for our work and we see the theoretical developments in here as key steps towards this direction, including the novel Theorem \ref{thm:smashing} which is a general result aimed at the problem of optimizing the reconstruction cost in a lattice based VAE.
  
Our work lies in the intersection of two streams of  work; the first exploiting ideas from information theory (and in particular rate distortion theory, which focuses on problem of lossy compression) in the context representation learning \cite{lindsay1983, tishby:bottleneck,tishby:dl_ib, DBLP:journals/corr/Shwartz-ZivT17,slonim_weiss_ml_ib, Watanabe2015, DBLP:journals/corr/GiraldoP13, kenneth_da, Banerjee:2004, clustering_bregman, higgins:beta, alemi:elbo, lastras:it_nll, NealHinton1998} , and the other devoted to the problem of learning discrete representations \cite{williams1992simple,DBLP:journals/corr/BengioLC13, mnih2014neural, schulman2015gradient, gu2015muprop, mnih2016variational, grathwohl2017backpropagation, maddison_concrete,  jang_gumbel, tucker_rebar, van2017neural, razavi2019generating}. We have chosen to contrast our work with two popular baselines - Concrete VAEs \cite{maddison_concrete, jang_gumbel} and Vector Quantization based VAEs (VQ-VAE). 
In what follows in our paper, we defined lattices formally, introduce an ELBO setup specifically designed for them, develop the ideas behind two training algorithms and conclude with our evaluation of their qualities and performance.

\captionsetup{belowskip=-7pt}
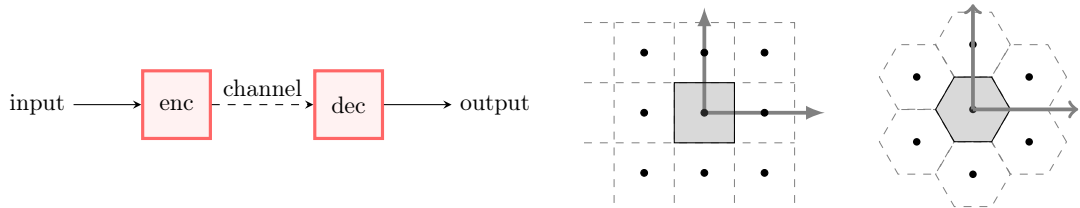
\begin{figure}
\begin{center}
\scalebox{1.0}{

\begin{minipage}{0.5 \textwidth}

\begin{tikzpicture}[scale=0.9,transform shape, x=0cm, y=0cm, >=stealth,squarednode/.style={rectangle, draw=red!60, fill=red!5, very thick, minimum size=10mm},]

\foreach \m/\l [count=\y] in {1,2,3,missing,4}
  \node [every neuron/.try, neuron \m/.try] (input-\m) at (0,2.5-\y) {};

\foreach \m [count=\y] in {1,missing,2}
  \node [every neuron/.try, neuron \m/.try ] (hidden-\m) at (2,2-\y*1.25) {};

\node (input)[]{input};
\node[squarednode]  (encoder) [right=of input] {enc};
\node[squarednode]  (decoder) [right=1.5cm of encoder] {dec};
\node (output)[right=of decoder]{output};
\draw[->,dashed](encoder)--++(decoder)    node[midway,above]  { channel };
\draw[->](input)--++(encoder);
\draw[->](decoder)--++(output);
\end{tikzpicture}

\end{minipage}

\begin{minipage}{0.25 \textwidth}
  \begin{tikzpicture}[scale=0.4]
    \coordinate (Origin)   at (0,0);
    \coordinate (XAxisMin) at (1,1);
    \coordinate (XAxisMax) at (5,1);
    \coordinate (YAxisMin) at (1,1);
    \coordinate (YAxisMax) at (1,4.5);
    \draw [thin, gray,-latex,line width=0.5mm] (XAxisMin) -- (XAxisMax);
    \draw [thin, gray,-latex,line width=0.5mm] (YAxisMin) -- (YAxisMax);

    \pgftransformcm{1}{0}{0}{1}{\pgfpoint{0cm}{0cm}}
    \coordinate (Bone) at (0,2);
    \coordinate (Btwo) at (2,-2);
    \draw[style=help lines,dashed] (-3,-2) grid[step=2cm] (4,4);
    \foreach \x in {-1,...,1}{
      \foreach \y in {-1,...,1}{
        \node[draw,circle,inner sep=0.9pt,fill] at (2*\x+1,2*\y+1) {};
      }
    }
    \filldraw[fill=gray, fill opacity=0.3, draw=black] (Origin)
        rectangle ($2*(Bone)+(Btwo)$);

  \end{tikzpicture}
  \end{minipage}

\begin{minipage}{0.25 \textwidth}
\newdimen\R
\R=1.2408cm
\begin{tikzpicture}[scale=0.4,darkstyle/.style={circle,draw,fill=gray!40,minimum size=20}]
  \foreach \x in {-1,...,1}
    \foreach \y in {-1,...,1} 
        	{
 
        	\ifthenelse{\cnttest{ (2*\x + \y)*(2*\x + \y)}{<}{9}}{

		\newcommand \ox { 1.732050807568877*1.078956521739130*\y};
		\newcommand \oy {2*1.078956521739130*\x + \y*1.078956521739130};
		

		\coordinate (offset) at (\ox,\oy); 
      		 \node [fill=black,circle,inner sep=0pt, minimum size=0.1cm]  (\x\y) at (offset) {}; 
		 
	 \draw[style=help lines,dashed] ++(\ox,\oy)  +(0:\R) \foreach \x in {60,120,...,360} {
                -- +(\x:\R)
            }   node[above] {} ;

	}{}
	}
	
	 \draw[fill] ++(0,0)  +(0:\R)[line width=0.075mm, fill=gray, fill opacity=0.3] \foreach \x in {60,120,...,360} { -- +(\x:\R) };

	\draw[thin,gray,->,line width=0.5mm] (0,0) -- (3.5,0);
	\draw[thin,gray ,->,line width=0.5mm] (0,0) -- (0,3.5);

\end{tikzpicture}
\end{minipage}
  
  }
  \end{center}
  \caption{A digital channel interpretation of a discrete representation and the $\mathbb{Z}^2$ and $A_2$ lattices. }
  \label{fig:lattice}
\end{figure}



\section{Lattices and a variant of variational inference}
A lattice is the set of all points that can be found by integral combinations of the row vectors of a matrix $B \in \realnumber^{m \times m}$, and where $B$ has full rank:
\begin{eqnarray*}
\Lambda(B) = \left\{ \mathbf{i} \cdot B   : \mathbf{i} \in \mathbb{Z}^m \right\}.
\end{eqnarray*}
For example, if $B = I_m$ we obtain the (product) $m$ dimensional integer lattice 
\begin{eqnarray*}
\mathbb{Z}^m = \{\cdots, -2,-1,0,1,2, \cdots \}^m.
\end{eqnarray*}
In this article, in addition to the integer lattice $\mathbb{Z}$ ($m=1$) we will also employ the square lattice $\mathbb{Z}^2$, the hexagonal lattice $A_2$ ($m=2$) with basis
\begin{eqnarray*}
\left[
\begin{array}{cc} 0 & 1 \\ 
\sqrt{3/4} & 1/2 \\ 
\end{array} \right]
\end{eqnarray*}
as well as the $E_8$ ($m=8$) lattice with basis
\begin{eqnarray*}
\left[
\begin{array}{cccccccc} 
2  &  0  & 0 & 0 & 0 & 0 & 0 & 0\\ 
-1 &  1  & 0 & 0 & 0 & 0 & 0 & 0\\ 
0  & -1  & 1 & 0 & 0 & 0 & 0 & 0\\ 
0  &  0  &-1 & 1 & 0 & 0 & 0 & 0\\ 
0  &  0  & 0  &-1 & 1 & 0 & 0 & 0\\ 
0 & 0  &  0  & 0  &-1 & 1 & 0 & 0\\ 
0 & 0 & 0  &  0  & 0  &-1 & 1 & 0\\ 
1/2 & 1/2 & 1/2 & 1/2 & 1/2 & 1/2 & 1/2 & 1/2 \\
\end{array} \right].
\end{eqnarray*}
These lattices have simple nearest neighbor algorithms  \cite{convay1982fast} and the best known Normalized Second Moment  for their given dimension \cite{conway1982voronoi}. We will introduce in Section \ref{s:gproxy} the notion of Normalized Second Moment; for now it suffices to state that it is a way to measure the \emph{covering efficiency} of a lattice quantizer, which relates to the number of lattice cells needed to fill a fixed volume in space for a given quantization error. A primary hypothesis in this article is that the overall performance of a lattice based VAE is related to such covering efficiency; this is further discussed in the Appendix's Subsection \ref{ss:motivating}.

To train lattice valued latent variable models,  we will introduce a  deviation to the classical variational inference setup. We introduce a ``helper'' continuous random vector $U \in \realnumber^m$ which is correlated with $Z$. Together, these two define our new latent space $(Z,U)$:
\begin{eqnarray*}
p_X(x) = E_{Z,U \sim p_{Z|U}f_U} \left[ p_{X|ZU}(x | Z, U) \right].
\end{eqnarray*}


 

As with classical variational inference, we assume that the observations $\{x_1,\cdots,x_n\}$ are independent, and thus we seek to optimize the maximum likelihood target
\begin{eqnarray*}
\min_{\theta} -\frac{1}{n} \sum_{i=1}^n \log E_{Z,U \sim p_{Z|U}(\cdot;\theta)f_{U}(\cdot;\theta) } \left[ p_{X|Z,U}(x_i | Z, U;\theta)) \right].
\end{eqnarray*}
To solve the optimization problem, we modify the ELBO to account for $U$ as seen below, where $q_{Z|X, U}$ is called the \emph{approximate posterior} or \emph{encoder}, and $p_{X|Z, U}$ is called the \emph{decoder}:
\begin{eqnarray*}
-\frac{1}{n} \sum_{i=1}^n \log E_{Z, U \sim p_{Z|U}f_U } \left[ p_{X|Z, U}(x_i | Z, U)) \right] = -\frac{1}{n} \sum_{i=1}^n \log E_{U \sim f_U} \left[ E_{Z  \sim p_{Z|U} } \left[ p_{X|Z, U}(x_i | Z, U) \right]  \right]  \\
\leq  -\frac{1}{n} \sum_{i=1}^n E_{U \sim f_U} \left[  E_{Z \sim q_{Z|X, U} (\cdot|x_i, U)} \left[ \log \left( \frac{p_{Z|U}(Z|U) }{q_{Z|X, U}(Z|x_i, U)}p_{X|Z,U}(x_i | Z, U) \right) \right] \right].
\end{eqnarray*} 
Given a vector $v \in \realnumber^m$, we denote by $K_{\Lambda}(v)$  the closest vector in the lattice $\Lambda$ as measured by the $L_2$ norm.  In our work, we will work with a very specific form for $q_{Z|X, U}$. We will be using a deterministic continuous encoder $\enc : \mathcal{X} \rightarrow \realnumber^m$. Then $q_{Z|X, U}(\cdot | x, u)$ will assign mass 1 to 
\begin{eqnarray*}
K_{\Lambda}(\enc(x)+u).
\end{eqnarray*}
We will justify this choice of $q_{Z|XU}$ in Section \ref{sec:training}. Following the literature in information theory and signal processing \cite{Zamir96onlattice, zamir2009lattices, zamir_nazer_kochman_bistritz_2014}, we will regard $u$ as a ``dither'' and the act of adding it to $\enc(x)$ as a ``dithering'' of the (deterministic) encoding of $x$. Using this encoder, we can simplify the ELBO for $x_i$ as follows:
\begin{eqnarray}
-E_{U \sim f_U}   \left[ \log \ p_{Z|U}( K_{\Lambda}(\enc(x_i)+U)|U) \right] - E_{U \sim f_U}   \left[ \log p_{X|Z,U}(x_i | K_{\Lambda}(\enc(x_i)+U), U))  \right].
\label{eq:simplified_elbo}
\end{eqnarray}
These two terms are the representation and reconstruction costs, each governed by the encoder and decoder, respectively. The key problem is clearly how we get rid of the quantization in the ELBO we formulated. In the following sections, we will introduce two ideas aimed at accomplishing this.



\section{Learning Lattice Representations}
\label{sec:training}

One way to get rid of the quantization in the ELBO (\ref{eq:simplified_elbo})  is to use a result from information theory called the "Crypto-Lemma" \cite{forney:shannon_wiener,Zamir96onlattice, zamir2009lattices, zamir_nazer_kochman_bistritz_2014}. To illustrate, let $U$ be uniformly distributed over $[-0.5,0.5]$, and let $K$ denote rounding to the nearest integer. The essential observation is that the distributions of $y-U$ and $K(y+U)-U$ are identical as illustrated next:
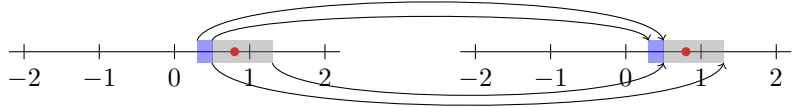
\begin{center}
\scalebox{1.0}{
  \begin{tikzpicture}
    \def \offset{-2}
    \draw (\offset-2.2,0) -- (\offset+2.2,0);
    \foreach \i in {-2,...,2} 
      \draw (\offset + \i,0.1) -- + ( 0,-0.2) node[below] {$\i$};
      \def \x{0.8};
      \fill[red] (\offset + \x,0) circle (0.6 mm);
      \filldraw [fill=gray, draw=none, fill opacity=0.4] ( \offset + \x + 0.5,0.15) rectangle (\offset + 0.5,-0.15);
      \filldraw [fill=blue, draw=none, fill opacity=0.4] (\offset + 0.5,0.15) rectangle ( \offset + \x - 0.5,-0.15);
      
          \def \offset{4}
    \draw (\offset-2.2,0) -- (\offset+2.2,0);
    \foreach \i in {-2,...,2} 
      \draw (\offset + \i,0.1) -- + ( 0,-0.2) node[below] {$\i$};
      \def \x{0.8};
      \fill[red] (\offset + \x,0) circle (0.6 mm);
      \filldraw [fill=gray, draw=none, fill opacity=0.4] ( \offset + \x + 0.5,0.15) rectangle (\offset + 0.5,-0.15);
      \filldraw [fill=blue, draw=none, fill opacity=0.4] (\offset + 0.5,0.15) rectangle ( \offset + \x - 0.5,-0.15);

      \draw[->] (\x-0.5 - 2,0.15)  to [out=70,in=110, looseness=0.3] (\offset + 0.5,0.15) ;
      \draw[->] (\x+0.5 - 2,-0.15)  to [out=-70,in=-110, looseness=0.3] (\offset + 0.5,-0.15) ;
      \draw[->] (0.5 - 2,-0.15)  to [out=-70,in=-110, looseness=0.3] (\offset + \x + 0.5,-0.15) ;

      \draw[->] (0.5 - 2,0.15)  to [out=70,in=110, looseness=0.2] (\offset + \x - 0.5,0.15) ;

  \end{tikzpicture}
  }
  \end{center}

The red dot represents $y$ and the blue/region in the left illustrates the uniform distribution $y+U$ (or $y-U$). Once you quantize, the blue region is mapped to 0, and the gray region is mapped to 1. On the right, we illustrate the distribution of $K(y+U)-U$. This is a much more general observation. Let $\mathcal{P}_0(\Lambda)$ denote the set of points in $\realnumber^m$ that are mapped to $0$ by $K_{\Lambda}$. Then we know the following:

\begin{lemma}[Crypto-Lemma \cite{zamir_nazer_kochman_bistritz_2014}]  Let $U$ be uniformly distributed over $\mathcal{P}_0(\Lambda)$. For any $y \in \realnumber^m$, the distribution of the random vector $K_{\Lambda}(y+U)-U$ is identical to the distribution of $y-U$.
\label{lem:crypto}
\end{lemma}
This Lemma will allow us to bridge between a continuous and discrete view of our VAE.

\subsection{The reconstruction cost term}
We  further simplify the type of decoder so that the dependency of $X$ on $Z$ and $U$ is solely through $Z-U$. Then using the Crypto Lemma, we obtain that for any choice of lattice $\Lambda$, and letting $U$ be uniformly distributed over $\mathcal{P}_0(\Lambda)$, the first term in (\ref{eq:simplified_elbo}) can be restated as
\begin{eqnarray*}
-E_{U \sim f_U}   \left[ \log p_{X|Z-U}(x_i | K_{\Lambda}(\enc(x_i)+U)-U)  \right] = - E_{U \sim f_U}   \left[ \log p_{X|Z-U}(x_i | \enc(x_i)-U))  \right].
\end{eqnarray*}
\subsection{The representation cost term}
Converting the representation cost term to one that does not use quantization is  significantly more challenging. The trick we employ relies on an auxiliary random vector $S \in \realnumber^m$ whose distribution ideally reflects the empirical distribution of the observed encodings $\{ \enc(x_1), \cdots \enc(x_n) \}$ as well as the corresponding unobserved (continuous) encodings, however obviously at training time there is no access to unobserved samples and furthermore, $\enc$ it self is being learned, making matters more difficult. We  assume that $S$ has some predefined prior distribution. We also assume $S$ is continuous and has a density, and quite importantly, that the corresponding density is positive when evaluated at any of the $\enc(x_i)$. This is our main result, which holds for any lattice $\Lambda$:

\begin{theorem} Assume $U$ is uniformly distributed over $\mathcal{P}_0$. Define $Z = K_{\Lambda}(S+U)$, and assume that $p_{Z|U}(z|u) > 0$ and $f_{S-U}(\eta) > 0$  for all $z \in \mathcal{Z}$, $u, \eta \in \mathbb{R}^m$. Then
\begin{eqnarray}
E_{U \sim f_U}   \left[ \log \frac{1}{\ p_{Z|U}( K_{\Lambda}(\enc(x_i)+U)|U)}\right]  = E_{U \sim f_U} \left[ \log \frac{
f_{U}(U)}{f_{S-U}( \enc(x_i) -U)} \right].
\label{eq:equiv}
\end{eqnarray}
\label{thm:smashing}
\end{theorem}
The proof uses  the Crypto-Lemma twice as well as Bayes' theorem. Let the notation $A \cisymbol B$ denote that $A$ is independent from $B$ and the notation $\ci{A}{B}{C}$ means that $A$ and $B$ are independent given $C$. We will use the following elementary Lemma:

\begin{lemma} Let $S, U, Z$ be $\Re^m$ valued random vectors such that $S \cisymbol U$ and $\ci{S}{U}{Z-U}$, with $U $ continuous and $Z$ discrete, and assume that $p_{Z|U}(z|u) > 0$  and $f_{Z-U}(\eta) > 0$ for any $z \in \mathcal{Z}$, $u, \eta \in \mathbb{R}^m$ .
Then
\begin{eqnarray*}
\frac{p_{Z|SU}(z|su)}{p_{Z|U}(z|u)} = \frac{f_{Z-U|S}(z-u|s)}{f_{Z-U}(z-u)}.
\end{eqnarray*}
\label{lem:useful}
\end{lemma}
\textbf{Proof of Lemma \ref{lem:useful}.} We assume in this proof that $S$ is continuous and has a density; the case where $S$ is discrete can be proved similarly. The proof proceeds as follows:
\begin{eqnarray*}
\frac{p_{Z|SU}(z|su)}{p_{Z|U}(z|u)} &\stackrel{(a)}{=}& \frac{f_{S|Z,U}(s|z,u)}{f_{S|U}(s|u)} \\
& \stackrel{(b)}{=}& \frac{f_{S|Z,U}(s|z,u)}{f_{S}(s)} \\
& \stackrel{(c)}{=}& \frac{f_{S|Z-U,U}(s|z-u,u)}{f_{S}(s)} \\
& \stackrel{(d)}{=}& \frac{f_{S|Z-U}(s|z-u)}{f_{S}(s)}  \\
& \stackrel{}{=}& \frac{f_{S|Z-U}(s|z-u)f_{Z-U}(z-u)}{f_{S}(s)f_{Z-U}(z-u)}  \\
& \stackrel{(e)}{=}& \frac{f_{Z-U|S}(z-u|s)}{f_{Z-U}(z-u)}  \\
\end{eqnarray*}
where $(a)$ follows from Bayes' rule for mixed continuous/discrete variables, $(b)$ follows from the assumption that $S \cisymbol U$, $(c)$ follows from the fact that these two events are identical
\begin{eqnarray*}
[Z=z, U=u] = [Z-U=z-u, U=u].
\end{eqnarray*}
Finally, $(d)$ follows from the assumption that $\ci{S}{U}{Z-U}$ and $(e)$ follows from the use of Bayes' rule.

\qed


\textbf{Proof of Theorem \ref{thm:smashing}.}   At the core of our proof is the classical argument employed in the analysis of lattices in quantization \cite{forney:shannon_wiener,Zamir96onlattice, zamir2009lattices, zamir_nazer_kochman_bistritz_2014}, which is that because by definition $U \in \mathcal{P}_0(\Lambda)$, 
\begin{eqnarray*}
K_{\Lambda}(Z -U) = Z
\end{eqnarray*}
and therefore $K_{\Lambda}(Z -U)  - (Z - U) = U$. This shows that $U$ can be obtained as function of $Z - U$, which implies that $\ci{S}{U}{Z-U}$ and  since by construction $S \cisymbol U$, we will be able to apply Lemma \ref{lem:useful}.


The proof thus proceeds as follows:
\begin{eqnarray*}
E_{U} \left[ \log \frac{1}{p_{Z|U}(K_{\Lambda}(e(x)+U)|U)} \right] &\stackrel{(a)}{=}& E_{U} \left[ \log \frac{p_{Z|S,U}( K_{\Lambda}( e(x) + U ) |e(x),U) }{p_{Z|U}(K_{\Lambda}(e(x)+U)|U)} \right] \\
&\stackrel{(b)}{=}& E_{U} \left[ \log \frac{f_{Z-U| S}(K_{\Lambda}( e(x) + U )-U|e(x))}{f_{Z-U}(K_{\Lambda}( e(x) + U )-U)} \right] \\
&\stackrel{(c)}{=}& E_{U} \left[ \log \frac{f_{Z-U| S }(e(x)-U|e(x))}{f_{Z-U}(e(x)-U)} \right] \\
&\stackrel{(d)}{=}& E_{U} \left[ \log \frac{f_{S-U|S}(e(x)-U|e(x))}{f_{S-U}(e(x)-U)} \right] \\
&\stackrel{(e)}{=}& E_{U} \left[ \log \frac{f_{U}(U)}{f_{S-U}(e(x)-U)} \right] \\
\end{eqnarray*}
where $(a)$ follows from the definition of $Z$ which implies that for all $x$ and all $u$,
\begin{eqnarray*}
p_{Z|U,S}(K_{\Lambda}(e(x)+u)|e(x),u) = 1,
\end{eqnarray*}
$(b)$ follows from an application of Lemma \ref{lem:useful}, $(c)$ follows from an application of the Crypto Lemma from which we know that the distributions of $K_{\Lambda}( e(x) + U )-U$ and $e(x)-U$ are identical, $(d)$ follows from another application of the Crypto Lemma, from which we can derive that $f_{Z-U} = f_{S-U}$ and $f_{Z-U|S} = f_{S-U|S}$ and $(e)$ follows from $S \cisymbol U$.
\qed

\subsection{The objective function }
\label{ss:objfun}
Putting together the representation and reconstruction cost terms, we obtain the training loss function
\begin{eqnarray}
L(x_i) = E_{U \sim f_U} \left[ \log \frac{
f_{U}(U)}{f_{S-U}( \enc(x_i) -U)} \right] - E_{U \sim f_U}   \left[ \log p_{X|Z-U}(x_i | \enc(x_i)-U) \right].
\label{eq:loss}
\end{eqnarray}
Note that if $U$ and $S$ were Gaussian, the above would correspond to a loss function that one would use when one trains a VAE with a Gaussian prior ($f_{S-U}$) and a Gaussian approximate posterior ($f_{U}$) with a data dependent mean and a data independent covariance matrix. For now,  ideally  we want to choose the distribution of $S$ so that $f_{S-U}$ and $p_{Z|U}$ have closed form expressions; note that $f_U(U)$ is easy to calculate from the basis $B$ using $f_U(U) = 1 / | \det(B) |$.  

One example where closed form expressions are feasible is the integer lattice and where $S$ is assumed to be a Laplacian distribution. We assume that the integer lattice is scaled using a parameter $\Delta$ so that the quantization points are $\{\cdots, -2 \Delta, -\Delta, 0, \Delta, 2 \Delta, \cdots\}$, and so that $U$ is uniformly distributed over $(-\Delta/2,\Delta/2)$.  Let $\alpha$ be the decay rate for the Laplacian distribution of $S$, then
\begin{eqnarray*}
\log \frac{f_U(u)}{f_{S-U}(s-u) }&=&  \left\{
\begin{array}{cc}
-\log(1 - \exp(-\alpha \Delta / 2 ) \cosh(\alpha (s-u)))  & | s-u | < \Delta/2  \\
\alpha | s-u | - \log( \sinh(\alpha \Delta/2))  &  | s-u | \geq \Delta/2 
\end{array}\right. , \\
\log \frac{1}{p_{Z|U}(z|u) }&=&  \left\{
\begin{array}{cc}
-\log(1 -  \exp( - \alpha \Delta / 2 ) \cosh(\alpha u) )& z = 0   \\
 \alpha | z \Delta - u | - \log( \sinh(\alpha \Delta/2))  &  z \neq 0 
\end{array}\right. .
\end{eqnarray*}

Implementing a VAE that can be trained using differentiation, yet used at inference time using quantization over the integer lattice $\mathbb{Z}$ is straightforward using (\ref{eq:loss}) and the equations above; our results, including Theorem \ref{thm:smashing}, assure us that in expectation, the performance of either will be identical. 
\subsection{A family of distributions over lattice points}
\label{ss:family}
For general choices for the lattice and $f_{S}$, closed form expressions for $f_{S-U}$ and $p_{Z|U}$ are harder or impossible to find. In the case of $f_{S-U}$, we can resort to sampling and averaging using $f_{S-U}(\eta) = k^{-1} \sum_{i=1}^k E_{V_i \sim f_U} \left[ f_{S}(\eta-V_i) \right]$. To work around the problem of $p_{Z|U}$, we will propose a simple class of parametric distributions over the lattice points $\Lambda$. To distinguish from $Z$ in Theorem \ref{thm:smashing}, we will denote any one instance of our proposed class  $p_{\hat{Z}|U}$. We then train using the composite loss 
\begin{eqnarray}
L(x_i) - E_{U \sim f_U}   \left[ \log  p_{\hat{Z}|U}( K_{\Lambda}(\enc(x_i)+U)|U) \right].
\label{eq:augloss}
\end{eqnarray}
The derivative of the second term with respect to any of the parameters of $\enc$ is zero from the standpoint of \texttt{autograd}. Thus the addition of this term leaves the optimization of the continuous deterministic encoder $\enc$ and the decoder $p_{X|Z-U}$ entirely governed by $L(x_i)$ as before.

The simplest useful distribution we could conceive of is one where probability of a lattice vector depends solely on its $L_2$ norm. We will be focused on \emph{integral lattices} \cite{conway2013sphere}, which have the property that  if $a,b \in \Lambda$, then the $a^Tb$ is an integer (and therefore, so is $\|a\|_2^2$), as is the case for the $\mathbb{Z}$ (and thus $\mathbb{Z}^2$), $A_2$ and $E_8$ lattices. The logarithm of the probability of a vector $a \in \Lambda$ will then be given by
\begin{eqnarray*}
\texttt{logsoftmax}(\Psi)_{ \| a \|_2^2 } - \log \theta_{ \| a \|_2^2} , 
\end{eqnarray*}
where $\Psi$ represents parametrized logits for the nonnegative integers and $\theta$ denotes the coefficients of the polynomial 
\begin{eqnarray*}
\Theta_{\Lambda}(\tau) = \sum_{x \in \Lambda}\exp(i\pi \tau \|x \|^2_2)
\end{eqnarray*}
called the theta function of a lattice; these coefficients are the number of lattice vectors of a given norm. We simply use a precomputed such list in practice \cite{sloane:oeis}. In our experiments, we scale lattices using a parameter $\Delta$ in which case the lattices are no longer integral; with proper de-scaling, it is elementary to continue to use the ideas above.

This simple distribution ignores any dependencies on $U$ for $p_{\hat{Z}|U}$ which can be important. In our experience, the most commonly used lattice vectors have a small norm, which merit learning a distribution for them that depends on $U$. Thus whenever $\|a\|_2^2 \geq t$ for some threshold $t$, we use the simple distribution described above, and whenever $\|a\|_2^2 < t$, we use a learned probability mass function that does depend on $U$. In addition, we learn a binary flag that denotes which of the two distributions is being used and we incorporate that in the overall representation cost.

\subsection{Training using a Gaussian proxy}
\label{s:gproxy}
There is a non-trivial sense in which $U$ approximates a Gaussian random vector as one increases the lattice dimension provided we use the right lattices and we already have acknowledged that (\ref{eq:loss}) has significant similarities with a Gaussian VAE's loss function. This observation  leads to another way of training lattice representations, based on using the encoder and decoder obtained when training a Gaussian VAE while separately learning the parameters of a distribution over the lattice points as in Subsection \ref{ss:family}. One advantage of this method is that a designer can simply focus on getting a good Gaussian VAE for a given problem, and then derive from this a lattice based VAE.

The \emph{volume}, \emph{second moment}  and \emph{normalized second moment} (NSM) of the lattice cell are defined as
\begin{eqnarray*}
V_{\Lambda} = \int_{\mathcal{P}_0(\Lambda)}  du, \hspace{0.25in} \sigma^2_{\Lambda} = \frac{1}{V_{\Lambda}} \frac{1}{m} \int_{\mathcal{P}_0(\Lambda)} \| u \|_2^2 du, \hspace{0.25in}  G(\Lambda) = \frac{\sigma^2_{\Lambda}}{V^{2/m}_{\Lambda}} 
\end{eqnarray*}
respectively. The smallest possible NSM for any lattice on dimension $m$ is defined as $G_m$.  The normalized second moment of an $m$-dimensional hyper-sphere is denoted by $G^*_m$. It is known that
\begin{eqnarray*}
G_m > G_m^* > \frac{1}{2 \pi e } \approx 0.058, \hspace{0.25in}   \lim_{m \rightarrow \infty} G_m = \frac{1}{2 \pi e };
\end{eqnarray*}
in other words, there exists a sequence of lattices $\Lambda_1, \Lambda_2, \cdots$ whose NSM approaches that of a hyper-sphere as the  lattice dimension $m$ grows to infinity.  For example, the NSMs for the $\mathbb{Z}$, $A_2$ and $E_8$ lattices are $0.0833, 0.0802$ and $0.0717$ respectively. Now denote $N^{*}_m$ to be an $m$ dimensional Gaussian vector with independent entries each with a variance equal to $\sigma^2_{\Lambda}$, and let $U_m$ denote a random vector uniformly distributed over $\mathcal{P}_{0}(\Lambda_{m})$. Then, it is not difficult to show that \cite{zamir_nazer_kochman_bistritz_2014}
\begin{eqnarray}
D_{KL}(U_m \| N^{*}_m ) = \frac{1}{2} \log (2 \pi  e G (\Lambda_m)), \hspace{0.25in}  \lim_{m \rightarrow \infty} D_{KL}(U_m \| N^{*} _m) = 0.
\label{eq:div}
\end{eqnarray}
where $D_{KL}$ denotes the Kullback-Liebler divergence \cite{kullback1951information, kullback1959information}.  These observations lead to the following idea:  suppose that we replace in Equation (\ref{eq:loss}) the random vectors $U$ and $S$ with zero mean Gaussian vectors $U_G$ and $S_G$ with (learned) covariance matrices $\sigma_{U_G}^2 I_m$ and $\sigma_{S_G}^2 I_m$, respectively, in effect training a Gaussian VAE. Next, for a given lattice $\Lambda$ with corresponding basis $B$, and a given scaling parameter $\Delta$, let $\Lambda_{\Delta}$ be the lattice associated with basis $\Delta B$. Next, we match $\Delta$ so that the second moment of $U$, a random vector drawn uniformly over $\Lambda_{\Delta}$, matches exactly that of $U_G$. It is easy to see that this can be accomplished by setting
 \begin{eqnarray}
\log \Delta =  \frac{1}{2} \log \sigma_{U_{G}}^2 - \frac{1}{2} \log G(\Lambda) - \frac{1}{m} \log V_{\Lambda}.
\label{eq:match_nsm}
\end{eqnarray}
Once it is time to do inference, we will ``forget'' that this was trained using Gaussian random variables, and instead dither the encoder output with $U$, apply lattice quantization, subtract the dither, and feed to the decoder. The idea is that matched second moments, together with the idea that good lattices have a noise distribution that eventually (as the dimension gets larger) resembles a Gaussian, results in an overall encoder/decoder network with similar behavior as that of the Gaussian one. 

To complete the discrete VAE, we learn a distribution  $p_{\hat{Z}|U}$ over the quantized output of the encoder $K_{\Lambda_{\Delta}}( e(x) + U )$. An example of such a distribution can be found in Subsection \ref{ss:family}. Learning the parameters of this distribution is straightforward: this can be done after the Gaussian VAE is trained, or at the same time that is being trained by using the trick in (\ref{eq:augloss}), which we do in our experiments.



\subsection{Summary of algorithms}
\label{ss:algo}
In the following we outline the algorithms we have discussed thus far. In here, $\Lambda$ is a given lattice, and $\enc$ and $-\log p_{X|Z-U}$ are parametrized encoder and decoder networks.

\subsubsection{Direct Lattice Training} 
The assumptions are that $f_S$ is a fixed distribution and  $\Delta$ is a free training parameter.
\begin{enumerate}
\item  Draw $U$ uniformly distributed over $\Lambda_{\Delta}$.
\item Compute $\enc(x)-U$, then feed to the decoder obtaining the reconstruction cost $-\log p_{X|Z-U}( x | \enc(x) - U)$. Compute also the KL term $\log f_{U}(U) / f_{S-U}(\enc(x)-U)$.
\item Perform a step of optimization on sum of the reconstruction and KL terms in step 2.
\end{enumerate}

\subsubsection{Training by Gaussian proxy} 

The assumptions are that $p_{\hat{Z}|U}$ is a given parametric form, and $\sigma_{U_G}^2, \sigma_{U_S}^2$ are training parameters.
\begin{enumerate}
\item Compute the matched lattice scaling factor $\Delta$ using (\ref{eq:match_nsm}). Draw $U$ uniformly distributed over $\Lambda_{\Delta}$, and $U_G  \sim \mathcal{N}(0, \sigma_{U_G}^2 I_m)$. 
\item Compute $\enc(x)-U_G$, then feed to the decoder obtaining the reconstruction cost 
\begin{eqnarray*}
-\log p_{X|Z-U}( x | \enc(x) - U_G). 
\end{eqnarray*}
Compute also the KL term 
\begin{eqnarray*}
\frac{1}{2}\left(  \frac{1}{\sigma_{U_G}^2 + \sigma_{U_S}^2}\| e(x) - U_G \|_2^2  - \frac{1}{ \sigma_{G}^2} \| U_G \|^2_2  - m \log \frac{\sigma_{U_G}^2}{\sigma_{U_G}^2 + \sigma_{U_S}^2} \right).
\end{eqnarray*}
\item Compute $K_{\Lambda_{\Delta}}(\enc(x) + U)$. Calculate the code length $-\log p_{\hat{Z}|U}( K_{\Lambda_{\Delta}}(\enc(x) + U)| U)$.
\item Optimization the sum of the reconstruction cost, the KL term and the code length.
\end{enumerate}
\subsubsection{Inference (for any training method)} 
\begin{enumerate}
\item Draw $U$ uniformly distributed over $\Lambda$.
\item Compute $ K_{\Lambda_{\Delta}}(\enc(x)+U) - U$, then feed to the decoder obtaining 
\begin{eqnarray}
-\log p_{X|Z-U}( x | K_{\Lambda_{\Delta}}(\enc(x)+U) - U).
\label{eq:reccost}
\end{eqnarray} 
\item Compute the code length 
\begin{eqnarray}
-\log p_{\hat{Z}|U}(K_{\Lambda_{\Delta}}(\enc(x)+U)|U).
\label{eq:repcost}
\end{eqnarray} 
\item The sum of (\ref{eq:reccost}) and (\ref{eq:repcost}) the (single sample) estimate for $p_X(x)$.
\end{enumerate}
The terminology "single sample" refers to the fact that the estimate is being calculated using a single $U$ sample. It is possible, and customary in the literature, to obtain improved estimates by using the idea of importance weighted autoencoders  \cite{DBLP:journals/corr/BurdaGS15}. This comes at the cost of increased complexity, proportional to the number of samples being used. 

\section{Evaluation}
We evaluate lattice based representations in the context of a problem of density estimation with latent variables, where we consider three baselines: Gaussian VAEs, Concrete VAEs, and VQ-VAEs, the latter using straight-through estimation \cite{DBLP:journals/corr/BengioLC13}. The first comparison we make is qualitative (Table \ref{tab:qualitative}). During inference time, our proposal is to do lattice quantization of the (dithered) output of an encoder, and to use a learned distribution over the lattice points as the prior in a latent variable model. This has many similarities with VQ-VAE, where the dictionary of vectors to which we could quantize is arbitrary; we could regard a lattice based system as a type of \emph{structured} VQ-VAE that has been further augmented with the idea of dithering, which allows us to make sharp statements about how it can be trained efficiently and its relation to Gaussian VAEs. In contrast, a Concrete VAE does not make a discrete representation explicit, as the output of the encoder is stochastic, and the corresponding representation cost is quantified using a KL divergence between an approximate posterior and a prior, instead of a direct probability mass function over the latent variables. 

\begin{table}
\caption{Qualitative comparison of discrete representations} 
\begin{center}
\begin{tabular}{cccc} \toprule
Discrete representation &  \begin{tabular}{@{}c@{}}Explicit discrete \\ representation \\\end{tabular}  &  \begin{tabular}{@{}c@{}} Gaussian model \\ link \\\end{tabular} &\begin{tabular}{@{}c@{}}Group\\ property \\\end{tabular} \\
Lattices &  Y & Y  & Y \\
Categorical+Concrete & N  & N  & N \\
VQ-VAE &  Y & N & N \\
\bottomrule
\end{tabular}
\end{center}
\label{tab:qualitative}
\end{table}

Next we observe that lattice based discrete representations provide a vehicle for transitioning from models involving Gaussian distributions to discrete representations, a property not shared with any other discrete representation that we know of. This is expected to be useful as statisticians may have significant experience designing and training models involving Gaussians. This is a leading motivation for introducing in Subsection \ref{s:gproxy} the idea of training by a Gaussian proxy.

Finally, we touch on the fact that due to their inherent nature, lattices may allow us to model the idea of \emph{composition} of representations, using the fact that lattices are a \emph{group}, this is, a set together with an operator which satisfy the axioms of closure, associativity, identity and invertibility. This property is unique to our proposal, and in principle could be exploited to build a simple formal algebra over representations of objects which can be learned through continuous differentiation methods. Demonstrating this idea is outside of the scope of this paper, but it is a key motivation for our work.


\subsection{Experiments}

In reference to the algorithms described in Subsection \ref{ss:algo}, our experiments report results on the direct lattice training algorithm for the $\mathbb{Z}$ lattice with  a Laplacian distribution for $S$ as described in Subsection \ref{ss:objfun}, and results on training by Gaussian proxy for the $\mathbb{Z}^2$, $A_2$, and $E_8$ lattices where in the latter we learn a probability mass function $p_{\hat{Z}|U}$ over the lattice vectors as described in Subsection \ref{ss:family}. It should be noted that the $\mathbb{Z}, A_2$ and $E_8$ lattices have the best known NSM for their respective dimensions. The main reason we use $\mathbb{Z}^2$ and not $\mathbb{Z}$ when training with Gaussian proxies is that for comparison purposes,  we wanted to use exactly the same implementation as with the $A_2$ lattice (simply changing the underlying lattice basis). Our implementation of both Concrete and VQ-VAE was reached after several iterations to improve their performance. Details are in the supporting documentation. Our code will be made publicly available.


Additionally, we use \emph{product lattices}. For a given lattice with dimension $m$ and basis $B$, we can create a (product) lattice with dimension $t$ (where $t$ is divisible by $m$) using the block diagonal basis $\mbox{diag}( \Delta_1 B, \Delta_2 B, \cdots, \Delta_{t/m} B)$ for scalar scaling parameters $\{ \Delta_i \}_{1}^{t/m}$. We scale the entire basis $B$ as opposed to each row vector individually because the nearest neighbor algorithm for the $E_8$ lattice applies only in the case the basis is scaled uniformly. In the case of the $\mathbb{Z}^2$, $A_2$, and $E_8$ lattices, the scaling parameter $\Delta_i$ is selected using (\ref{eq:match_nsm}); for the $\mathbb{Z}$+Laplace setting, it is learned directly. 

\begin{table}
\setlength{\tabcolsep}{4pt}
\caption{Test negative log likelihood for MNIST and OMNIGLOT}
\begin{tabular}{cccccccccc}
\toprule mnist & & \multicolumn{4}{c}{linear}& \multicolumn{4}{c}{gated nonlinear}\\ 
\cmidrule(r){3-6} \cmidrule(r){7-10} 
samples  & repr. & 16 & 24 & 32 & 200 & 16 & 24 & 32 & 200 \\\toprule 
1  & $\mathbb{Z}^2$ & 127.61 & 120.49 & 118.38 & 125.64 & 101.21 & 100.33 & 100.37 & 101.02 \\ 
  & $\mathbb{Z}$+laplace & 127.40 & 120.17 & 117.85 & 125.35 & 100.82 & 100.66 & 100.82 & 108.59 \\ 
  & $A_2$ & 127.24 & 119.87 & 117.61 & 123.87 & 100.66 & 99.77 & 99.87 & 100.70 \\ 
  & $E_8$ & 126.91 & 119.43 & 117.07 & 121.15 & \textbf{100.32} & \textbf{99.72} & \framebox{\textbf{99.66}} & \textbf{100.09} \\ 
  & VQ-VAE & \textbf{126.52} & \textbf{117.50} & \textbf{111.55} & \framebox{\textbf{107.21}} & 103.33 & 103.66 & 103.48 & 103.77 \\ 
\rowcolor{lightgray}  & concrete & 137.53 & 127.08 & 121.23 & 114.86 & 119.55 & 113.10 & 110.16 & 105.39 \\ 
\rowcolor{lightgray}  & gaussian & 124.54 & 116.00 & 112.47 & 109.48 & 98.01 & 95.93 & 95.80 & 96.12 \\ 
10K & $\mathbb{Z}^2$ & 123.05 & 114.55 & 111.45 & 114.41 & 95.60 & 94.38 & 94.35 & 94.92 \\ 
  & $\mathbb{Z}$+laplace & 122.99 & 114.31 & 111.15 & 116.29 & 95.62 & 94.99 & 95.26 & 101.74 \\ 
  & $A_2$ & 122.76 & 114.07 & 110.91 & 113.13 & \textbf{95.03} & 94.01 & 93.93 & 94.59 \\ 
  & $E_8$ & \textbf{122.22} & \textbf{113.19} & \textbf{109.86} & 110.39 & 95.29 & \textbf{93.55} & \framebox{\textbf{93.55}} & \textbf{93.66} \\ 
  & VQ-VAE & 126.52 & 117.50 & 111.55 & \framebox{\textbf{107.21}} & 103.33 & 103.66 & 103.48 & 103.77 \\ 
\rowcolor{lightgray}  & concrete & 135.20 & 123.57 & 116.78 & 107.17 & 115.67 & 107.19 & 101.86 & 94.65 \\ 
\rowcolor{lightgray}  & gaussian & 119.69 & 109.95 & 106.00 & 102.48 & 91.88 & 89.55 & 89.69 & 90.05 \\ 
\bottomrule 
 omniglot & & \multicolumn{4}{c}{linear}& \multicolumn{4}{c}{gated nonlinear}\\ 
\cmidrule(r){3-6} \cmidrule(r){7-10} 
samples &repr.& 16 & 24 & 32 & 200 & 16 & 24 & 32 & 200 \\\toprule 
 1 & $\mathbb{Z}^2$ & 145.54 & 141.99 & 139.94 & 146.33 & 132.95 & 129.16 & 125.79 & 128.95 \\ 
  & $\mathbb{Z}$+laplace & 145.63 & 141.96 & 139.94 & 146.14 & 132.74 & 128.43 & 128.45 & 134.64 \\ 
  & $A_2$ & 145.10 & \textbf{141.22} & 139.31 & 144.21 & 133.15 & 127.80 & 125.16 & 128.94 \\ 
  & $E_8$ & \textbf{144.85} & 141.24 & 138.81 & 140.94 & \textbf{131.26} & \textbf{127.67} & \framebox{\textbf{124.42}} & \textbf{126.31} \\ 
  & VQ-VAE & 148.06 & 141.36 & \textbf{136.85} & \framebox{\textbf{126.17}} & 135.72 & 132.05 & 130.72 & 134.21 \\ 
\rowcolor{lightgray}  & concrete & 141.40 & 134.88 & 130.24 & 121.06 & 143.06 & 136.34 & 132.63 & 122.21 \\ 
\rowcolor{lightgray}  & gaussian & 142.38 & 137.34 & 133.93 & 127.35 & 130.52 & 123.82 & 120.21 & 120.33 \\ 
10K & $\mathbb{Z}^2$ & 141.23 & 136.10 & 133.10 & 133.82 & 125.00 & 120.08 & 117.60 & 118.83 \\ 
  & $\mathbb{Z}$+laplace & 141.44 & 136.25 & 133.22 & 136.53 & 124.84 & 120.30 & 119.72 & 125.24 \\ 
  & $A_2$ & 140.88 & 135.64 & 132.66 & 132.42 & 124.79 & 119.34 & 117.15 & 118.73 \\ 
  & $E_8$ & \textbf{140.37} & \textbf{134.89} & \textbf{131.47} & 129.14 & \textbf{123.57} & \textbf{118.75} & \framebox{\textbf{115.75}} & \textbf{116.33} \\ 
  & VQ-VAE & 148.06 & 141.36 & 136.85 & \framebox{\textbf{126.17}} & 135.72 & 132.05 & 130.72 & 134.21 \\ 
\rowcolor{lightgray}  & concrete & 140.40 & 133.47 & 128.39 & 117.38 & 137.51 & 128.69 & 123.09 & 112.38 \\ 
\rowcolor{lightgray}  & gaussian & 138.20 & 131.75 & 127.81 & 120.53 & 119.56 & 114.21 & 111.29 & 111.62 \\ 
\bottomrule 
\end{tabular}
\label{tab:all}
\end{table}

We consider two architectures where encoder and decoder share a similar structure, with the exception of the input and output dimensions; here we describe the encoders: 1) A simple fully connected linear model $z = xA  + b$  and 2) A two deterministic layer network where the first layer is a gated nonlinear network and the second layer is a fully connected network: $z = \texttt{relu}(\sigma(x G + b_0) * (x A_1 + b_1)) A_2 + b_2$, where $\sigma$ denotes the sigmoid function used for gating, and $*$ denotes element-wise product; the former is very common in VAE studies and the latter is an architecture we found in \cite{DBLP:journals/corr/TomczakW17}.

We will report results for 112 different configurations, varying the hidden size length $(16,24,32,200)$, the network architecture (linear and gated nonlinear), the hidden layer type ($\mathbb{Z}^2$, $\mathbb{Z}$+Laplace, $A_2$, $E_8$, VQ-VAE, Concrete and Gaussian), and the data set (MNIST \cite{staticmnist} and OMNIGLOT \cite{omniglot}). For every single configuration, we performed 30 experiments with different random seeds, leading to a total of 33690 experiments. Seven different learning rates are scanned twice (1e-4, 3e-4, 5e-4, 7e-4, 1e-3, 3e-3, 5e-3). After those 14 experiments, the learning rate with the best validation value is selected and used for the remaining 16. From the set of 30 experiments, the one with the best validation is chosen, and then we report on the test set performance for a single test sample or 10k test samples. We use Adam \cite{kingma:adam} as the optimizer, together with a learning rate annealing schedule that halves the learning rate after the validation loss does not show an improvement for more than 50 epochs. After 200 epochs with no improvement on the validation loss, we stop the experiment. There is no limit on the total number of epochs. Concrete's implementation follows \cite{tucker_rebar} with a temperature of 0.1, whereas VQ-VAE's follows \cite{van2017neural}, concatenating 4 codes each with 512 vectors with the trick of adding to the loss function a term to force the encoder output to approximate the quantized output (weight factor 1/4); we scanned the options of 2 or 8 codes (with 512 vectors) or 256 and 1024 vectors (4 codes) before settling on the above. To document the computational complexity of our proposed method in its various configurations, we compare in the supporting material the empirical epoch duration times and total number of epochs with those of other baselines. The MNIST dataset we use is a commonly used static binarized version of the original dataset. The OMNIGLOT validation and test datasets are binarized by drawing Bernoulli random bits using a bias given by the original image; we use the original OMNIGLOT (non-binarized) dataset during training. The NSMs for the $\mathbb{Z}$, $A_2$ and $E_8$ lattices are $0.0833, 0.0802$ and $0.0717$ respectively, the best known for their respective dimensions.

The experimental results are summarized in Table \ref{tab:all}, which is meant to illustrate two possible extremes using importance sampling \cite{DBLP:journals/corr/BurdaGS15}. To create a data compression system which compresses the latent variable first, and then compresses the data conditional on the latent variable, then test samples = 1 applies, whereas if the goal is to perform density estimation and we can tolerate high computational costs, the relevant results are for when test samples = 10K.  The best result for each hidden size/test sampling/data set/model combination is highlighted, and if one optimizes over the chosen hidden layer sizes, then the best result is doubly highlighted. When highlighting, neither Gaussian nor Concrete are considered, because the former is not discrete and because the latter is not an explicit discrete representation. 
We summarize below our main experimental conclusions:
\begin{enumerate}[leftmargin=*]
\item A lattice with a lower NSM leads, generally, to a better log likelihood, as we hoped. This adds evidence to the idea that lattices can approximate Gaussian performance. The differences in performance are quite small nonetheless which required very careful experimental setup; statistics on the distributions of the performance are in the supporting material.
\item Product lattices perform better for smaller hidden layer sizes than for larger ones. This is likely because of inefficiencies that are inherent in lower dimensional lattices which do not disappear when used in a product lattice construction. Our best option to overcome this is to use lattices with lower NSM (and hence higher dimensional), but good constructions with simple nearest neighbor algorithms are not elementary to find.
\item Lattice based methods improve with more test samples (as does Concrete and Gaussian) because even though they are explicit representations, they are still conditioned on randomness that helps during importance sampling. In contrast, VQ-VAE does not improve with additional importance samples because of its reliance on deterministic quantization.
\item There is no basis to prefer VQ-VAE to lattices (or vice versa) based on this experiment for the simple tasks considered herein, as their relative competitiveness appears to be dependent on the network architecture. If one wants the "group property", then lattices are the only alternative.
\item To our surprise, Gaussian is not always better than Concrete,  calling into question the wisdom of assuming that the ultimate goal is to approximate a Gaussian VAE, as lattices do. Researching finite alphabet coding techniques for approximating Concrete performance is thus interesting.
\end{enumerate}


\section{Summary}
In this article we introduced the idea of using lattices as a representation space in latent variable generative models, motivated by the fact that lattice based VAEs give us mechanisms for constructing explicit discrete representations with  connections to Gaussian VAEs, and also the \emph{group property}, which we intend to exploit in subsequent research to create formal algebras on latent variables that can be learned through differentiation methods. For a given variational inference loss function that employs dithered lattice quantization, we demonstrated that we can find an equivalent one that does not employ quantization by leveraging a new theorem targeting the representation cost in a lattice based variational auto encoder. Using these ideas, we derived the equations for simple integer lattice VAE, introduced ideas for training more general lattices and further specialized these to the case of training by Gaussian proxies which leverage the connection between lattice and Gaussian VAEs. Our experimental results suggest lattices can be competitive even without accounting for their special properties which are of independent interest. Our immediate next step is to study how to take advantage of the group property of lattice representations.

\section{Appendix}

 \subsection{Stochastic, deterministic and explicit representations}
\label{ss:digression}

In the main article, when we introduced the ELBO
\begin{eqnarray}
\log p_X(x) \geq E_{Z \sim Q_{Z|X}(\cdot | x)} \left[ \log \frac{p_Z(Z)}{Q_{Z|X}(Z|x)} \right] + E_{Z \sim Q_{Z|X}(\cdot | x)} \left[ \log p_{X|Z}(x | Z) \right].
\label{eq:elbo}
\end{eqnarray}
we remarked that $Q_{Z|X}$, in its most general form provides a stochastic representation of the object $x$. The idea that the optimum such encoder is in general stochastic was surprising when it was first discovered, and it led to an proposal, called ``bits-back-coding'' \cite{hinton1993keeping} to interpret  the first term in the expression above (a Kullback-Liebler divergence) as type of code, leading to the so-called information theoretic interpretation of a VAE.

The bits back coding argument is probably best discussed in the context of a data compression application. The idea here is   to send $z$ at the cost of $-\log p_Z(z)$ and then $x$ at the cost of $-\log p_{X|Z}(x|z)$. Sending $z$ and then $x$ this way nonetheless in general exceeds the cost that the ELBO gives. The bits back argument is that you can ``get back'' some of the bits that you spent this way because you could cleverly hide a message in the choice of $z$. 

A more straightforward situation (which need not rely on the bits back argument) arises if one can choose $Q_{Z|X}$ to be deterministic, because then the representation cost term collapses to a simple code length:
\begin{eqnarray*}
- E_{Z \sim Q_{Z|X}(\cdot | x)} \left[ \log p_{Z}(Z) \right]
\end{eqnarray*}
The reader may object nonetheless - didn't we just state that in general the optimum solution is stochastic? Yes, in general it is! However, the experiments in the original VQ-VAE proposal (and for that matter, even our own experiments) show that the performance from such deterministic choices may still be good. We believe that there are good theoretical explanations for this phenomenon which can be traced back all the way back to Shannon's original coding theorems  \cite{shannon1948mathematical, Sha59}, which connect expressions involving scalar random variables with deterministic high dimensional codes; this is left for a future exposition.

In our proposal, lattice based VAEs are not strictly speaking deterministic because representations are \emph{conditional} on a dither $U$ which is assumed to be shared randomness between encoder and decoder:
\begin{eqnarray}
-E_{U \sim f_U}   \left[ \log \ p_{Z|U}( K_{\Lambda}(\enc(x_i)+U)|U) \right] - E_{U \sim f_U}   \left[ \log p_{X|Z,U}(x_i | K_{\Lambda}(\enc(x_i)+U), U))  \right].
\label{eq:simplified_elbo}
\end{eqnarray}
Having said this, lattice based representations are certainly not stochastic in the original VAE sense - the reader can see in the expression above that the representation $K_{\Lambda}(\enc(x_i)+U)$ is very much deterministic given $U$ and that the cost of the representation is measured using a straight conditional code length  $-\log  p_{Z|U}$. Thus in lattice VAEs, as in VQ-VAE, one need not invoke the bits-back argument  to give an interpretation to the ELBO as a coding system; in either of these two the ELBO has a much more straightforward interpretation as a two stage coding system. For this reason, we refer to both VQ-VAE and lattice based representations as \emph{explicit}.

Concrete based VAE	s, although undeniably allowing us to construct a latent variable model with discrete random variables, train an encoder that is \emph{not} explicit in general and therefore it becomes harder to directly use it as a means of building a data compression system. It's quite possible, nonetheless, that we maybe able to find more explicit constructions of discrete representations based on the Concrete distribution idea. In this case, instead of relying on lattices on Euclidean spaces we likely want to focus our attention on finite alphabet coding theory, and in particular, the theory of linear algebraic codes. This is left for future research.

\subsection{Motivating the use of covering efficient lattices}
\label{ss:motivating}
An important claim in our article is that more efficient lattices, in the sense of their Normalized Second Moment (NSM), can lead to better overall log likelihood performance in a discrete lattice variable latent model.  In this section we discuss more in detail why this is a reasonable claim to make.

As a reminder, the computational network we are experimenting with uses a continuous encoder $e(x)$, a dither $U$, a quantizer $K_{\Lambda}$, a representation cost distribution $p_{Z|U}$ and a decoder $p_{X|Z-U}$ assembled together in the following end to end expression:
\begin{eqnarray*}
-\log p_{X|Z-U}( x | K_{\Lambda}(e(x+U)) - U ) - \log p_{Z|U}( K_{\Lambda}(e(x+U))  | U ) 
\end{eqnarray*}
As with all variational autoencoders, there is an inherent tradeoff between the two terms above. If the lattice $\Lambda$ (equivalently $K_{\Lambda}$) introduces little quantization error, then the term in the left, \emph{allegedly}, produces better estimates of the probability of $x$ as the information about $x$ has not been ``corrupted much''. Nonetheless, if $K_{\Lambda}$ introduces little quantization error, then its output has a higher descriptive complexity (as one varies the input $x$) than compared to the output of a lattice which is allowed to introduce higher quantization error. The descriptive complexity is measured by the $-\log p_{Z|U}$ term (the pointwise conditional Shannon entropy).

One way to picture the tradeoff between descriptive complexity and the error introduced by lattice quantization is to imagine that for a \emph{fixed volume}, we are filling the space in this volume with lattice cells associated with a given error (second moment):
\begin{eqnarray*}
\sigma^2_{\Lambda} = \frac{1}{V_{\Lambda}} \frac{1}{m} \int_{\mathcal{P}_0(\Lambda)} \| u \|_2^2 du 
\end{eqnarray*}

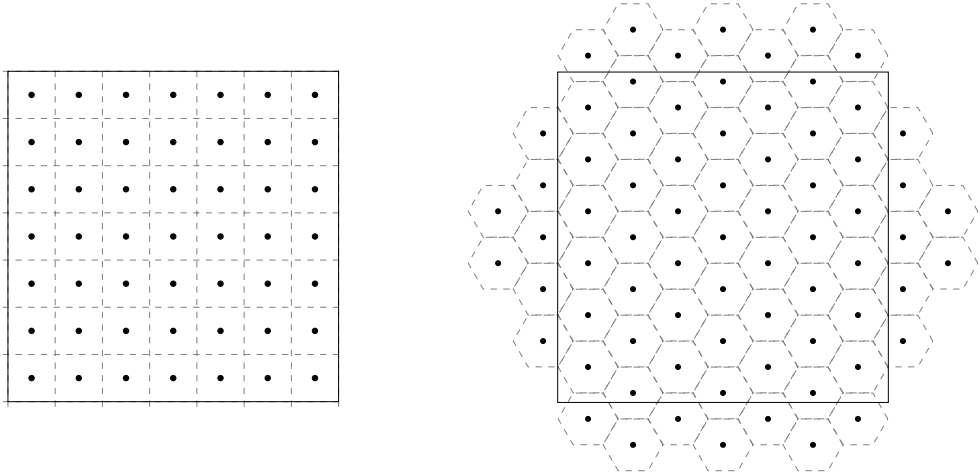
\begin{figure}
\begin{center}
\scalebox{0.8}{
	\FPeval{\boxlen}{1.9624}
	\FPeval{\halfboxlen}{\boxlen/2}
\begin{minipage}{0.5 \textwidth}
  \begin{tikzpicture}[scale=0.4]
    \coordinate (Origin)   at (0,0);
    \coordinate (XAxisMin) at (1,1);
    \coordinate (XAxisMax) at (5,1);
    \coordinate (YAxisMin) at (1,1);
    \coordinate (YAxisMax) at (1,4.5);

    \pgftransformcm{1}{0}{0}{1}{\pgfpoint{0cm}{0cm}}
    \coordinate (Bone) at (0,\boxlen);
    \coordinate (Btwo) at (\boxlen,-\boxlen);
    \draw[style=help lines,dashed,xshift=-\halfboxlen cm,yshift=- \halfboxlen cm] (-6.1,-6.1) grid[step=\boxlen cm] (8,8);
    \foreach \x in {-3,...,3}{
      \foreach \y in {-3,...,3}{
        \node[draw,circle,inner sep=0.9pt,fill] at (1.9624*\x,1.9624*\y) {};
      }
    }

\FPeval{\rectlen}{\boxlen*3.5};

\draw (-\rectlen,-\rectlen) -- (\rectlen,-\rectlen) -- (\rectlen,\rectlen) -- (-\rectlen,\rectlen) -- cycle;

  \end{tikzpicture}
  \end{minipage}
\begin{minipage}{0.5 \textwidth}
\newdimen\R
\R=1.2408cm
\begin{tikzpicture}[scale=0.4,darkstyle/.style={circle,draw,fill=gray!40,minimum size=20}]
  \foreach \x in {-5,...,5}  {
    	\foreach \y in {-5,...,5}  {
		\newdimen \ox
		\newdimen \oy
		\ifthenelse{\cnttest{3*\y*\y+(2*\x+\y)*(2*\x+\y)}{<}{80}}{

		\FPeval{\ox}{1.732050807568877*1.078956521739130*\y};
		\FPeval{\oy}{2*1.078956521739130*\x + \y*1.078956521739130};
			\coordinate (offset) at (\ox,\oy); 
      			\node [fill=black,circle,inner sep=0pt, minimum size=0.1cm]  (\x\y) at (offset) {}; 
		 	\draw[style=help lines,dashed] ++(\ox,\oy)  +(0:\R) \foreach \x in {60,120,...,360} {
				-- +(\x:\R)
			}   node[above] {} ;    
		}
	}
 }

\FPeval{\rectlen}{\boxlen*3.5};

\draw (-\rectlen,-\rectlen) -- (\rectlen,-\rectlen) -- (\rectlen,\rectlen) -- (-\rectlen,\rectlen) -- cycle;

\end{tikzpicture}
\end{minipage}}
  \caption{The $\mathbb{Z}^2$ and $A_2$ lattices used to cover the same area. The drawing has been adjusted so that the second moment of both lattice cells is the same. }
  \label{fig:lattice}
  \end{center}
\end{figure}

where
\begin{eqnarray*}
V_{\Lambda} = \int_{\mathcal{P}_0(\Lambda)}  du, \hspace{0.25in} 
\end{eqnarray*}
and $\mathcal{P}_0(\Lambda)$ denotes the lattice cell that contains the origin. The descriptive complexity is smallest when the number of cells needed to fill this volume is the smallest.

In Figure \ref{fig:lattice} we show a comparison of two lattices as they are used to fill the same volume (area in this case, as we are operating in two dimensions). The first lattice is the $\mathbb{Z}^2$ lattice, and the second lattice is $A_2$. The dimensions of the drawings have been adjusted so that the second moment of both the square and hexagonal lattice cells are identical, and the square bold boxes in either of the two drawings are identical; from the definition fo the Normalized Second Moment
\begin{eqnarray*}
G(\Lambda) = \frac{\sigma^2_{\Lambda}}{V^{2/m}_{\Lambda}}
\end{eqnarray*}
and using the fact that in this case $m=2$, then we obtain that
\begin{eqnarray*}
G(\Lambda_{\mathbb{Z}^2}) V_{\Lambda_{\mathbb{Z}^2}} = \sigma^2_{\Lambda_{\mathbb{Z}^2}} = \sigma^2_{\Lambda_{A_2}} = G(\Lambda_{A_2}) V_{\Lambda_{A_2}} 
\end{eqnarray*}
and therefore 
\begin{eqnarray*}
\frac{V_{\Lambda_{\mathbb{Z}^2}}}{V_{\Lambda_{A_2}} } =  \frac{G(\Lambda_{\mathbb{Z}^2})}{ G(\Lambda_{A_2})} \approx 0.963...
\end{eqnarray*}
which implies that filling a large area would use at least 3 \% fewer lattice  points with the hexagonal lattice than with the square lattice. In Figure \ref{fig:lattice} the area in the left contains 49 lattice points whereas the one in the right contains 45 lattice points. There are nonetheless some unaccounted for sections of lattices in the borders, as the expectation is that it would take approximately 47.1... lattice cells to fill the same area. 

As the reader may appreciate, the difference between the space filling efficiency of these two lattices is quite small, which created significant challenges in our experimental setup; this is one of the reasons we found it necessary to attempt a number of experiments from random seeds (30) per configuration that is higher than what can be normally found in the literature. Still, the reader may notice in the experimental results in the main paper that indeed, the hexagonal lattice system does do slightly better than the square one (here we are comparing the $\mathbb{Z}^2$ and the $A_2$ results, not with the Laplace + $\mathbb{Z}$ since the latter uses a different prior than the first two).

Having stated this, our goal is, naturally, to approach Gaussian performance using practical lattice based techniques. The present article does \emph{not} accomplish this goal. The closest we got was to propose the use of the $E_8$ lattice.

The $E_8$ lattice is the subset of $\mathbb{R}^8$ comprised of all the integers with the following two properties:
\begin{itemize}
\item Either all coordinates are integers, or all are half integers.
\item The sum of all coordinates is even.
\end{itemize}
Furthermore, a simple exact nearest neighbor algorithm is known  \cite{convay1982fast}. Let $f(x)$ denote the act of rounding to the nearest integer all the entries of $x$ and let $g(x)$ denote the vector where the entry that would cause the largest error after rounding, is rounded the opposite (wrong) way. Then the nearest neighbor algorithm is as follows: compute $f(x), g(x), f(x-0.5)+0.5, g(x-0.5)+0.5$., and choose the one that is closest to $x$.

At the present moment, the $E_8$ lattice has the best known Normalized Second Moment amongst all lattices in 8 dimensions. As it can be seen from the results in the main paper, we were able to show that using the $E_8$ lattice further improves log likelihood performance beyond that of the hexagonal $A_2$ lattice.

To gain additional performance, higher dimensional lattices are called for. One possibility is to use the 24 dimensional Leech lattice \cite{leech1967notes, conway1982voronoi, conway2013sphere}, which can be constructed using the $E_8$ lattice; nearest neighbor algorithms for this lattice have been proposed \cite{adoul1988nearest}.

\section{Additional Experimental statistics}

\subsection{Experimental resolution}
One key question is whether the experimental setup we have conceived has sufficient resolution to assert our claim that improving the Normalized Second Moment can improve the overall VAE performance.

In Tables \ref{tab:mean_std_mnist} and \ref{tab:mean_std_omniglot} we report mean and standard deviation for all the experiments that we did for any particular configuration. The standard deviations shown are often in the order of the gaps separating techniques that are very close in performance (but for which we claim we have statistical evidence of them being different). Having said this, the average performance is not as important as the best performance, and therefore, to establish our claim it is necessary to rely on other ways of analyzing the data. 

What we decided to do is to compute the probability that particular type of intermediate representation is better than another one when sampling at random from their respective experiments (each with different random seeds and potential different learning rates), and comparing the results.  The corresponding results can be found in Table \ref{tab:comparison}.

The way to interpret these results is that they are answering the question: is a representation in a row better than a representation in a column? If the corresponding entry in the table is larger than $1/2$, it implies than more than $1/2$ of the time the answer to this question is "yes", and therefore a number close to $1$ suggests confidence that this is indeed the case.

\begin{table}
\begin{center}
\caption{Competitive comparison for hidden dimension 32}
\begin{tabular}{ccccccccc}
\toprule 
MNIST, linear  & $\mathbb{Z}^2$ & $\mathbb{Z}$+Laplace& $A_2$& $E_8$& VQ-VAE& Concrete& Gaussian\\ 
 \toprule 
$\mathbb{Z}^2$ & 0.48& 0.00& 0.00& 0.00& 0.00& 1.00& 0.00\\ 
$\mathbb{Z}$+Laplace & 1.00& 0.48& 0.00& 0.00& 0.00& 1.00& 0.00\\ 
$A_2$ & 1.00& 1.00& 0.48& 0.01& 0.00& 1.00& 0.00\\ 
$E_8$ & 1.00& 1.00& 0.99& 0.48& 0.00& 1.00& 0.00\\ 
VQ-VAE & 1.00& 1.00& 1.00& 1.00& 0.48& 1.00& 0.04\\ 
Concrete & 0.00& 0.00& 0.00& 0.00& 0.00& 0.48& 0.00\\ 
Gaussian & 1.00& 1.00& 1.00& 1.00& 0.96& 1.00& 0.48\\ 
\bottomrule 
\end{tabular}
\begin{tabular}{ccccccccc}
\toprule 
 MNIST, gated & $\mathbb{Z}^2$ & $\mathbb{Z}$+Laplace& $A_2$& $E_8$& VQ-VAE& Concrete& Gaussian\\ 
 \toprule 
$\mathbb{Z}^2$ & 0.48& 0.82& 0.08& 0.03& 1.00& 1.00& 0.00\\ 
$\mathbb{Z}$+Laplace & 0.18& 0.48& 0.07& 0.01& 1.00& 1.00& 0.00\\ 
$A_2$ & 0.92& 0.93& 0.48& 0.29& 1.00& 1.00& 0.00\\ 
$E_8$ & 0.97& 0.99& 0.71& 0.48& 1.00& 1.00& 0.00\\ 
VQ-VAE & 0.00& 0.00& 0.00& 0.00& 0.48& 1.00& 0.00\\ 
Concrete & 0.00& 0.00& 0.00& 0.00& 0.00& 0.48& 0.00\\ 
Gaussian & 1.00& 1.00& 1.00& 1.00& 1.00& 1.00& 0.48\\ 
\bottomrule 
\end{tabular}
\begin{tabular}{ccccccccc}
\toprule 
OMNIGLOT, linear & $\mathbb{Z}^2$ & $\mathbb{Z}$+Laplace& $A_2$& $E_8$& VQ-VAE& Concrete& Gaussian\\ 
 \toprule 
$\mathbb{Z}^2$ & 0.48& 0.63& 0.00& 0.00& 0.00& 0.00& 0.00\\ 
$\mathbb{Z}$+Laplace & 0.37& 0.48& 0.00& 0.00& 0.00& 0.00& 0.00\\ 
$A_2$ & 1.00& 1.00& 0.48& 0.29& 0.00& 0.00& 0.00\\ 
$E_8$ & 1.00& 1.00& 0.71& 0.48& 0.01& 0.00& 0.00\\ 
VQ-VAE & 1.00& 1.00& 1.00& 0.99& 0.48& 0.00& 0.00\\ 
Concrete & 1.00& 1.00& 1.00& 1.00& 1.00& 0.48& 1.00\\ 
Gaussian & 1.00& 1.00& 1.00& 1.00& 1.00& 0.00& 0.48\\ 
\bottomrule 
\end{tabular}
\begin{tabular}{ccccccccc}
\toprule 
 OMNIGLOT, gated & $\mathbb{Z}^2$ & $\mathbb{Z}$+Laplace& $A_2$& $E_8$& VQ-VAE& Concrete& Gaussian\\ 
 \toprule 
$\mathbb{Z}^2$ & 0.48& 1.00& 0.04& 0.00& 1.00& 1.00& 0.00\\ 
$\mathbb{Z}$+Laplace & 0.00& 0.48& 0.00& 0.00& 0.88& 0.85& 0.00\\ 
$A_2$ & 0.96& 1.00& 0.48& 0.04& 1.00& 1.00& 0.00\\ 
$E_8$ & 1.00& 1.00& 0.96& 0.48& 1.00& 1.00& 0.00\\ 
VQ-VAE & 0.00& 0.12& 0.00& 0.00& 0.48& 0.64& 0.00\\ 
Concrete & 0.00& 0.15& 0.00& 0.00& 0.36& 0.48& 0.00\\ 
Gaussian & 1.00& 1.00& 1.00& 1.00& 1.00& 1.00& 0.48\\ 
\bottomrule 
\end{tabular}
\label{tab:comparison}
\end{center}
\end{table}

Examination of these tables results, in our opinion, in an endorsement of the main claims of the paper: improving the Normalized Second Moment (excluding $\mathbb{Z}$+Laplace, which uses a different training algorithm and prior), results in a better result, and furthermore, VQ-VAE and the $E_8$ switch their place as the best system depending on the type of model being used. It must be noted that while VQ-VAE and Concrete typically obtained even better results for the hidden size 200, the lattice based methods did not (and in fact they became worse), and therefore in principle these tables do not reflect the entire story; still we believe they do bring confidence that the comparisons done in the main paper have validity.

\subsection{Epoch Elapsed times}

Our experiments used V100 GPUs running on either x86 or PowerPC hosts. In Table \ref{tab:all} we document the elapsed times, in milliseconds, for one training epoch for the OMNIGLOT and gated nonlinear model. It can be seen that the fastest algorithms are VQ-VAE, Gaussian, and the Laplace+$\mathbb{Z}$ lattice based VAE but at the same time, the lattice based methods are quite practical.

\begin{table}
\begin{center}
\setlength{\tabcolsep}{4pt}
\caption{Elapsed time in milliseconds for OMNIGLOT and the gated nonlinear model}
\begin{tabular}{cccc}
\toprule omniglot  & \multicolumn{3}{c}{gated nonlinear 1s 1d}\\ 
\cmidrule(r){2-4} 
 \begin{tabular}{@{}c@{}}hidden \\repr. \\\end{tabular} & 16 & 24 & 32 \\\toprule
   $\mathbb{Z}^2$ & 1580.62 & 1926.99 & 1973.80 \\ 
   $\mathbb{Z}$+Laplace & 1059.85 & 1109.95 & 1278.91 \\ 
   $A_2$ & 1509.57 & 2034.65 & 2077.38 \\ 
   $E_8$ & 1985.97 & 2367.38 & 2476.59 \\ 
   VQ-VAE & 1029.14 & 889.03 & 927.82 \\ 
 Concrete & 1405.04 & 1198.54 & 1371.35 \\ 
  Gaussian & 869.81 & 1417.53 & 1266.50 \\ 
\bottomrule 
\end{tabular}
\label{tab:all}
\vspace{-0.1in}
\end{center}
\end{table}

\section{Additional implementation details}

\subsection{Simulating a random variable uniformly distributed over a lattice cell}

Let $V$ be a random row $m$ vector with independent entries uniformly distributed over $[0,1]$ and define
\begin{eqnarray*}
W = V B
\end{eqnarray*}
A random variable uniformly distributed over the lattice cell is then given by $W-K_{\Lambda}( W ) \in \mathcal{P}_0(\Lambda)$.

\subsection{Our implementation of Concrete}
Our first attempt at implementing Concrete VAEs followed that of the original article \cite{maddison_concrete}, but we encountered difficulties in obtaining good performance. We settled on an idea found in a tensorflow implementation of \cite{tucker_rebar} (\url{https://github.com/tensorflow/models/tree/master/research/rebar}), (see also \cite{jang_gumbel}). The idea is to use, during training, an expression that is not guaranteed to be a lower bound on the log likelihood, yet in practice, appears to work well. During inference time, a proper lower bound is used. The details can be found in our implementation, released with this article.

\begin{table}
\begin{center}%
\caption{Mean and standard deviation for the test  NLL for MNIST}
\begin{tabular}{cccccc}
\toprule mnist & & \multicolumn{4}{c}{linear 1s}\\ 
\cmidrule(r){3-6} 
\begin{tabular}{@{}c@{}}test \\samples \\\end{tabular} & \begin{tabular}{@{}c@{}}hidden \\repr. \\\end{tabular} & 16 & 24 & 32 & 200 \\\toprule1 & baseline &   &   &   \\ 
  & $\mathbb{Z}^2$ & 127.74/0.16 & 120.62/0.12 & 118.43/0.04 & 125.87/0.16 \\ 
  & $\mathbb{Z}$+Laplace & 127.51/0.13 & 120.27/0.23 & 118.04/0.11 & 125.50/0.09 \\ 
  & $A_2$ & 127.33/0.15 & 120.03/0.17 & 117.63/0.07 & 124.23/0.14 \\ 
  & $E_8$ & 127.04/0.21 & 119.55/0.27 & 117.14/0.11 & 121.41/0.19 \\ 
  & VQ-VAE & 126.77/0.30 & 118.48/0.54 & 113.75/0.72 & 107.74/0.73 \\ 
\rowcolor{lightgray}  & Concrete & 138.70/0.67 & 128.19/0.85 & 122.02/0.43 & 116.03/0.62 \\ 
\rowcolor{lightgray}  & Gaussian & 124.64/0.13 & 116.02/0.11 & 112.41/0.04 & 109.48/0.05 \\ 
10K & $\mathbb{Z}^2$ & 123.15/0.13 & 114.58/0.11 & 111.50/0.04 & 114.60/0.12 \\ 
  & $\mathbb{Z}$+Laplace & 123.10/0.12 & 114.42/0.17 & 111.27/0.10 & 116.36/0.05 \\ 
  & $A_2$ & 122.84/0.13 & 114.15/0.14 & 110.93/0.05 & 113.40/0.11 \\ 
  & $E_8$ & 122.36/0.20 & 113.29/0.22 & 109.90/0.08 & 110.57/0.12 \\ 
  & VQ-VAE & 126.77/0.30 & 118.48/0.54 & 113.75/0.72 & 107.74/0.73 \\ 
\rowcolor{lightgray}  & Concrete & 136.01/0.43 & 124.37/0.47 & 117.31/0.29 & 107.91/0.40 \\ 
\rowcolor{lightgray}  & Gaussian & 119.74/0.10 & 110.00/0.07 & 106.02/0.03 & 102.53/0.02 \\ 
\bottomrule 
\end{tabular}
\begin{tabular}{cccccc}
\toprule mnist & & \multicolumn{4}{c}{gated nonlinear 1s 1d}\\ 
\cmidrule(r){3-6} 
\begin{tabular}{@{}c@{}}test \\samples \\\end{tabular} & \begin{tabular}{@{}c@{}}hidden \\repr. \\\end{tabular} & 16 & 24 & 32 & 200 \\\toprule1 & baseline &   &   &   \\ 
  & $\mathbb{Z}^2$ & 101.47/0.21 & 100.76/0.27 & 100.79/0.33 & 101.10/0.08 \\ 
  & $\mathbb{Z}$+Laplace & 101.21/0.19 & 100.86/0.16 & 101.05/0.22 & 108.88/0.24 \\ 
  & $A_2$ & 101.11/0.22 & 100.20/0.32 & 100.21/0.39 & 100.73/0.02 \\ 
  & $E_8$ & 100.99/0.22 & 100.03/0.26 & 100.02/0.25 & 100.67/0.75 \\ 
  & VQ-VAE & 104.12/0.40 & 104.48/0.42 & 104.48/0.58 & 105.79/1.40 \\ 
\rowcolor{lightgray}  & Concrete & 121.01/0.53 & 114.30/0.54 & 111.48/0.44 & 106.68/0.74 \\ 
\rowcolor{lightgray}  & Gaussian & 98.34/0.15 & 96.25/0.20 & 96.11/0.23 & 96.36/0.12 \\ 
10K & $\mathbb{Z}^2$ & 95.88/0.40 & 94.79/0.35 & 94.73/0.33 & 94.93/0.01 \\ 
  & $\mathbb{Z}$+Laplace & 96.14/0.35 & 95.42/0.29 & 95.40/0.31 & 102.13/0.34 \\ 
  & $A_2$ & 95.57/0.39 & 94.34/0.40 & 94.28/0.39 & 94.59/0.01 \\ 
  & $E_8$ & 95.70/0.28 & 93.86/0.32 & 93.84/0.27 & 94.16/0.60 \\ 
  & VQ-VAE & 104.12/0.40 & 104.48/0.42 & 104.48/0.58 & 105.79/1.40 \\ 
\rowcolor{lightgray}  & Concrete & 115.98/0.53 & 107.21/0.59 & 102.60/0.45 & 95.46/0.46 \\ 
\rowcolor{lightgray}  & Gaussian & 92.00/0.29 & 89.89/0.30 & 89.94/0.28 & 90.25/0.19 \\ 
\bottomrule 
\end{tabular}
\label{tab:mean_std_mnist}
\end{center}
\end{table}

\begin{table}
\begin{center}
\caption{Mean and standard deviation for the test  NLL for OMNIGLOT}
\begin{tabular}{cccccc}
\toprule omniglot & & \multicolumn{4}{c}{linear 1s}\\ 
\cmidrule(r){3-6} 
\begin{tabular}{@{}c@{}}test \\samples \\\end{tabular} & \begin{tabular}{@{}c@{}}hidden \\repr. \\\end{tabular} & 16 & 24 & 32 & 200 \\\toprule1 & baseline &   &   &   \\ 
  & $\mathbb{Z}^2$ & 145.51/0.08 & 141.93/0.12 & 140.00/0.06 & 146.50/0.12 \\ 
  & $\mathbb{Z}$+Laplace & 145.58/0.08 & 141.94/0.11 & 140.04/0.10 & 146.01/0.09 \\ 
  & $A_2$ & 145.15/0.11 & 141.17/0.07 & 139.19/0.11 & 144.60/0.21 \\ 
  & $E_8$ & 144.93/0.14 & 141.16/0.10 & 139.08/0.15 & 141.32/0.19 \\ 
  & VQ-VAE & 148.03/0.06 & 141.25/0.08 & 138.49/0.44 & 127.59/1.07 \\ 
\rowcolor{lightgray}  & Concrete & 141.62/0.19 & 135.11/0.15 & 130.60/0.20 & 121.78/0.55 \\ 
\rowcolor{lightgray}  & Gaussian & 142.42/0.10 & 137.28/0.10 & 133.97/0.07 & 127.46/0.09 \\ 
10K & $\mathbb{Z}^2$ & 141.25/0.05 & 136.12/0.04 & 133.18/0.07 & 133.96/0.09 \\ 
  & $\mathbb{Z}$+Laplace & 141.37/0.05 & 136.23/0.04 & 133.27/0.05 & 136.47/0.07 \\ 
  & $A_2$ & 140.92/0.05 & 135.68/0.06 & 132.62/0.05 & 132.61/0.12 \\ 
  & $E_8$ & 140.36/0.06 & 134.88/0.04 & 131.64/0.10 & 129.39/0.15 \\ 
  & VQ-VAE & 148.03/0.06 & 141.25/0.08 & 138.49/0.44 & 127.59/1.07 \\ 
\rowcolor{lightgray}  & Concrete & 140.60/0.15 & 133.64/0.10 & 128.64/0.17 & 117.88/0.40 \\ 
\rowcolor{lightgray}  & Gaussian & 138.10/0.11 & 131.76/0.05 & 127.87/0.05 & 120.57/0.03 \\ 
\bottomrule 
\end{tabular}
\begin{tabular}{cccccc}
\toprule omniglot & & \multicolumn{4}{c}{gated nonlinear 1s 1d}\\ 
\cmidrule(r){3-6} 
\begin{tabular}{@{}c@{}}test \\samples \\\end{tabular} & \begin{tabular}{@{}c@{}}hidden \\repr. \\\end{tabular} & 16 & 24 & 32 & 200 \\\toprule1 & baseline &   &   &   \\ 
  & $\mathbb{Z}^2$ & 132.75/0.65 & 129.05/0.53 & 126.03/0.30 & 129.74/0.79 \\ 
  & $\mathbb{Z}$+Laplace & 133.33/0.73 & 146.09/40.24 & 129.47/1.39 & 134.33/0.53 \\ 
  & $A_2$ & 132.52/0.61 & 128.06/0.37 & 125.20/0.34 & 129.03/0.09 \\ 
  & $E_8$ & 131.55/0.61 & 127.84/0.46 & 124.46/0.21 & 128.64/2.06 \\ 
  & VQ-VAE & 135.26/0.31 & 133.03/0.64 & 131.52/1.13 & 134.94/1.74 \\ 
\rowcolor{lightgray}  & Concrete & 142.66/1.21 & 137.04/1.40 & 131.87/1.65 & 122.73/0.71 \\ 
\rowcolor{lightgray}  & Gaussian & 129.83/0.71 & 123.89/0.41 & 120.06/0.33 & 120.24/0.60 \\ 
10K & $\mathbb{Z}^2$ & 124.91/0.19 & 120.18/0.14 & 117.83/0.35 & 119.42/0.59 \\ 
  & $\mathbb{Z}$+Laplace & 125.43/0.32 & 138.50/42.06 & 121.39/0.89 & 125.49/0.36 \\ 
  & $A_2$ & 124.64/0.23 & 119.67/0.14 & 117.25/0.24 & 118.95/0.21 \\ 
  & $E_8$ & 123.88/0.24 & 118.91/0.16 & 115.97/0.28 & 117.91/1.39 \\ 
  & VQ-VAE & 135.26/0.31 & 133.03/0.64 & 131.52/1.13 & 134.94/1.74 \\ 
\rowcolor{lightgray}  & Concrete & 137.16/1.19 & 129.49/1.16 & 122.89/0.84 & 112.78/0.47 \\ 
\rowcolor{lightgray}  & Gaussian & 119.54/0.23 & 114.25/0.11 & 111.62/0.35 & 111.79/0.37 \\ 
\bottomrule 
\end{tabular}
\label{tab:mean_std_omniglot}
\end{center}
\end{table}



\small

\bibliography{references}

\begin{thebibliography}{10}

\bibitem{peterson_anderson:mean_field_1987}
Carsten Peterson and James~R. Anderson.
\newblock {A mean field theory learning algorithm for neural networks }.
\newblock {\em Complex Systems}, 1(5):995--1019, 1987.

\bibitem{parisi:statistical_field_theory}
Giorgio Parisi.
\newblock {\em Statistical Field Theory}.
\newblock Addison-Wesley, 1988.

\bibitem{saul1996exploiting}
Lawrence~K. Saul and Michael~I. Jordan.
\newblock Exploiting tractable substructures in intractable networks.
\newblock In {\em Advances in neural information processing systems}, pages
  486--492, 1996.

\bibitem{saul1996mean}
Lawrence~K. Saul, Tommi Jaakkola, and Michael~I. Jordan.
\newblock Mean field theory for sigmoid belief networks.
\newblock {\em Journal of artificial intelligence research}, 4:61--76, 1996.

\bibitem{jaakkola1997variational}
Tommi Jaakkola and Michael~I. Jordan.
\newblock A variational approach to bayesian logistic regression models and
  their extensions.
\newblock In {\em Sixth International Workshop on Artificial Intelligence and
  Statistics}, volume~82, 1997.

\bibitem{ghahramani1997factorial}
Z~Ghahramani and MI~Jordan.
\newblock Factorial hidden markov models machine learning.
\newblock {\em Kluwer Academic Publishers}, 1997.

\bibitem{jordan1999introduction}
Michael~I Jordan, Zoubin Ghahramani, Tommi~S Jaakkola, and Lawrence~K Saul.
\newblock An introduction to variational methods for graphical models.
\newblock {\em Machine learning}, 37(2):183--233, 1999.

\bibitem{hinton1993keeping}
Geoffrey~E Hinton and Drew Van~Camp.
\newblock Keeping the neural networks simple by minimizing the description
  length of the weights.
\newblock In {\em Proceedings of the sixth annual conference on Computational
  learning theory}, pages 5--13, 1993.

\bibitem{neal1998view}
Radford~M Neal and Geoffrey~E Hinton.
\newblock A view of the em algorithm that justifies incremental, sparse, and
  other variants.
\newblock In {\em Learning in graphical models}, pages 355--368. Springer,
  1998.

\bibitem{dempster1977maximum}
Arthur~P Dempster, Nan~M Laird, and Donald~B Rubin.
\newblock Maximum likelihood from incomplete data via the em algorithm.
\newblock {\em Journal of the Royal Statistical Society: Series B
  (Methodological)}, 39(1):1--22, 1977.

\bibitem{kingma:VAE}
D.P. Kingma and M.~Welling.
\newblock {Auto-Encoding Variational Bayes}.
\newblock In {\em The International Conference on Learning Representations
  (ICLR), Banff}. 2014.

\bibitem{shannon1948mathematical}
Claude~E Shannon.
\newblock A mathematical theory of communication.
\newblock {\em Bell system technical journal}, 27(3):379--423, 1948.

\bibitem{Sha59}
C.~E. Shannon.
\newblock {Coding theorems for a discrete source with a fidelity criterion}.
\newblock In {\em IRE Nat. Conv. Rec., Pt. 4}, pages 142--163. 1959.

\bibitem{berger1971rate}
T.~Berger.
\newblock {\em Rate Distortion Theory: A Mathematical Basis for Data
  Compression}.
\newblock Prentice-Hall electrical engineering series. Prentice-Hall, 1971.

\bibitem{van2017neural}
Aaron van~den Oord, Oriol Vinyals, et~al.
\newblock Neural discrete representation learning.
\newblock In {\em Advances in Neural Information Processing Systems}, pages
  6306--6315, 2017.

\bibitem{conway2013sphere}
John~Horton Conway and Neil James~Alexander Sloane.
\newblock {\em Sphere packings, lattices and groups}, volume 290.
\newblock Springer Science \& Business Media, 2013.

\bibitem{lindsay1983}
Bruce~G. Lindsay.
\newblock The geometry of mixture likelihoods: A general theory.
\newblock {\em Ann. Statist.}, 11(1):86--94, 03 1983.

\bibitem{tishby:bottleneck}
N.~Tishby, F.~Pereira, and W.~Bialek.
\newblock {The information bottleneck method}.
\newblock In {\em Proceedings of the 37-th Annual Allerton Conference on
  Communication, Control and Computing}, pages 368--377. 1999.

\bibitem{tishby:dl_ib}
Naftali Tishby and Noga Zaslavsky.
\newblock Deep learning and the information bottleneck principle.
\newblock In {\em 2015 IEEE Information Theory Workshop}. 2015.

\bibitem{DBLP:journals/corr/Shwartz-ZivT17}
Ravid Shwartz{-}Ziv and Naftali Tishby.
\newblock Opening the black box of deep neural networks via information.
\newblock {\em CoRR}, abs/1703.00810, 2017.

\bibitem{slonim_weiss_ml_ib}
Noam Slonim and Yair Weiss.
\newblock Maximum likelihood and the information bottleneck.
\newblock In {\em Proceedings of the 15th International Conference on Neural
  Information Processing Systems}, NIPS'02, pages 351--358, Cambridge, MA, USA,
  2002. MIT Press.

\bibitem{Watanabe2015}
Kazuho Watanabe and Shiro Ikeda.
\newblock Entropic risk minimization for nonparametric estimation of mixing
  distributions.
\newblock {\em Machine Learning}, 99(1):119--136, Apr 2015.

\bibitem{DBLP:journals/corr/GiraldoP13}
Luis Gonzalo~S{\'{a}}nchez Giraldo and Jos{\'{e}}~C. Pr{\'{\i}}ncipe.
\newblock Rate-distortion auto-encoders.
\newblock {\em CoRR}, abs/1312.7381, 2013.

\bibitem{kenneth_da}
Kenneth Rose.
\newblock Deterministic annealing for clustering, compression, classification,
  regression, and related optimization problems.
\newblock {\em Proceedings of the IEEE}, (11):2210--2239, November 1998.

\bibitem{Banerjee:2004}
Arindam Banerjee, Inderjit Dhillon, Joydeep Ghosh, and Srujana Merugu.
\newblock An information theoretic analysis of maximum likelihood mixture
  estimation for exponential families.
\newblock In {\em Proceedings of the Twenty-first International Conference on
  Machine Learning}, ICML '04, pages 8--, New York, NY, USA, 2004. ACM.

\bibitem{clustering_bregman}
Arindam Banerjee, Srujana Merugu, Inderjit Dhillon, and Joydeep Ghosh.
\newblock The geometry of mixture likelihoods: A general theory.
\newblock {\em Journal of Machine Learning Research}, (6):1705--1749, 2005.

\bibitem{higgins:beta}
I.~Higgins, L.~Matthey, A.~Pal, C.~Burgess, X.~Glorot, M.~Botvinick,
  S.~Mohamed, and A.~Lerchner.
\newblock {Beta-VAE: Learning Basic Visual Concepts with a Constrained
  Variational Framework.}
\newblock In {\em The International Conference on Learning Representations
  (ICLR), Toulon}. 2017.

\bibitem{alemi:elbo}
Alexander~A. Alemi, Ben Poole, Ian Fischer, Joshua~V. Dillon, Rif~A. Saurous,
  and Kevin Murphy.
\newblock Fixing a broken elbo.
\newblock In {\em Proceedings of the 35th International Conference on Machine
  Learning}. 2018.

\bibitem{lastras:it_nll}
Luis~A. Lastras-Monta{\~n}o.
\newblock Information theoretic lower bounds on negative log likelihood.
\newblock In {\em The International Conference on Learning Representations
  (ICLR), Toulon}. 2019.

\bibitem{NealHinton1998}
Radford~M. Neal and Geoffrey~E. Hinton.
\newblock {\em A View of the Em Algorithm that Justifies Incremental, Sparse,
  and other Variants}, pages 355--368.
\newblock Springer Netherlands, Dordrecht, 1998.

\bibitem{williams1992simple}
Ronald~J Williams.
\newblock Simple statistical gradient-following algorithms for connectionist
  reinforcement learning.
\newblock {\em Machine learning}, 8(3-4):229--256, 1992.

\bibitem{DBLP:journals/corr/BengioLC13}
Yoshua Bengio, Nicholas L{\'{e}}onard, and Aaron~C. Courville.
\newblock Estimating or propagating gradients through stochastic neurons for
  conditional computation.
\newblock {\em CoRR}, abs/1308.3432, 2013.

\bibitem{mnih2014neural}
Andriy Mnih and Karol Gregor.
\newblock Neural variational inference and learning in belief networks.
\newblock {\em arXiv preprint arXiv:1402.0030}, 2014.

\bibitem{schulman2015gradient}
John Schulman, Nicolas Heess, Theophane Weber, and Pieter Abbeel.
\newblock Gradient estimation using stochastic computation graphs.
\newblock In {\em Advances in Neural Information Processing Systems}, pages
  3528--3536, 2015.

\bibitem{gu2015muprop}
Shixiang Gu, Sergey Levine, Ilya Sutskever, and Andriy Mnih.
\newblock Muprop: Unbiased backpropagation for stochastic neural networks.
\newblock {\em arXiv preprint arXiv:1511.05176}, 2015.

\bibitem{mnih2016variational}
Andriy Mnih and Danilo~J Rezende.
\newblock Variational inference for monte carlo objectives.
\newblock {\em arXiv preprint arXiv:1602.06725}, 2016.

\bibitem{grathwohl2017backpropagation}
Will Grathwohl, Dami Choi, Yuhuai Wu, Geoffrey Roeder, and David Duvenaud.
\newblock Backpropagation through the void: Optimizing control variates for
  black-box gradient estimation.
\newblock {\em arXiv preprint arXiv:1711.00123}, 2017.

\bibitem{maddison_concrete}
Chris~J. Maddison, Andriy Mnih, and Yee~Whye Teh.
\newblock The concrete distribution: {A} continuous relaxation of discrete
  random variables.
\newblock In {\em 5th International Conference on Learning Representations
  (ICLR 2017)}. 2017.

\bibitem{jang_gumbel}
Eric Jang, Shixiang Gu, and Ben Poole.
\newblock Categorical reparametrization with gumbel-softmax.
\newblock In {\em 5th International Conference on Learning Representations
  (ICLR 2017)}. 2017.

\bibitem{tucker_rebar}
George Tucker, Andriy Mnih, Chris~J Maddison, John Lawson, and Jascha
  Sohl-Dickstein.
\newblock Rebar: Low-variance, unbiased gradient estimates for discrete latent
  variable models.
\newblock In {\em Advances in Neural Information Processing Systems 30}, pages
  2627--2636. Curran Associates, Inc., 2017.

\bibitem{razavi2019generating}
Ali Razavi, Aaron van~den Oord, and Oriol Vinyals.
\newblock Generating diverse high-fidelity images with vq-vae-2.
\newblock In {\em Advances in Neural Information Processing Systems}, pages
  14837--14847, 2019.

\bibitem{convay1982fast}
JH~Convay and NJA Sloane.
\newblock Fast quantizing and decoding algorithms for lattice quantizers.
\newblock {\em IEEE Trans Inform Theory}, 28(2):227--232, 1982.

\bibitem{conway1982voronoi}
J~Conway and N~Sloane.
\newblock Voronoi regions of lattices, second moments of polytopes, and
  quantization.
\newblock {\em IEEE Transactions on Information Theory}, 28(2):211--226, 1982.

\bibitem{Zamir96onlattice}
Ram Zamir and Meir Feder.
\newblock On lattice quantization noise.
\newblock {\em IEEE Trans. Inform. Theory}, 42:1152--1159, 1996.

\bibitem{zamir2009lattices}
Ram Zamir.
\newblock Lattices are everywhere.
\newblock In {\em 2009 Information Theory and Applications Workshop}, pages
  392--421. IEEE, 2009.

\bibitem{zamir_nazer_kochman_bistritz_2014}
Ram Zamir, Bobak Nazer, Yuval Kochman, and Ilai Bistritz.
\newblock {\em Lattice Coding for Signals and Networks: A Structured Coding
  Approach to Quantization, Modulation and Multiuser Information Theory}.
\newblock Cambridge University Press, 2014.

\bibitem{forney:shannon_wiener}
G.~D. Forney~Jr.
\newblock {Shannon meets Wiener II: On MMSE estimation in successive decoding
  schemes}.
\newblock In {\em In Proceedings of 42st Annual Allerton Conference on
  Communication, Control, and Computing}, 2004.

\bibitem{sloane:oeis}
Neil Sloane.
\newblock The on-line encyclopedia of integer sequences.
\newblock \url{https://oeis.org/}, 1964.

\bibitem{kullback1951information}
Solomon Kullback and Richard~A Leibler.
\newblock On information and sufficiency.
\newblock {\em The annals of mathematical statistics}, 22(1):79--86, 1951.

\bibitem{kullback1959information}
Solomon Kullback.
\newblock Information theory and statistics. john riley and sons.
\newblock {\em Inc. New York}, 1959.

\bibitem{DBLP:journals/corr/BurdaGS15}
Yuri Burda, Roger~B. Grosse, and Ruslan Salakhutdinov.
\newblock {Importance Weighted Autoencoders}.
\newblock In {\em {The International Conference on Learning Representations
  (ICLR)}}. 2016.

\bibitem{DBLP:journals/corr/TomczakW17}
Jakub~M. Tomczak and Max Welling.
\newblock {VAE} with a {VampPrior}.
\newblock In {\em The 21nd International Conference on Artificial Intelligence
  and Statistics}. 2018.

\bibitem{staticmnist}
Hugo Larochelle and Iain Murray.
\newblock The neural autoregressive distribution estimator.
\newblock In {\em Proceedings of the 14th International Conference on
  Artificial Intelligence and Statistics (AISTATS)}. 2011.

\bibitem{omniglot}
Brenden~M. Lake, Ruslan Salakhutdinov, and Joshua~B. Tenenbaum.
\newblock Human-level concept learning through probabilistic program induction.
\newblock In {\em Science}, volume 350, page 1332?1338. 2015.

\bibitem{kingma:adam}
Diederik~P. Kingma and Jimmy Ba.
\newblock {Adam: A Method for Stochastic Optimization}.
\newblock In {\em Proceedings of the 3rd International Conference on Learning
  Representations (ICLR)}. 2015.

\bibitem{leech1967notes}
John Leech.
\newblock Notes on sphere packings.
\newblock {\em Canadian Journal of Mathematics}, 19:251--267, 1967.

\bibitem{adoul1988nearest}
J-P Adoul and Michel Barth.
\newblock Nearest neighbor algorithm for spherical codes from the leech
  lattice.
\newblock {\em IEEE transactions on information theory}, 34(5):1188--1202,
  1988.

\end{thebibliography}
\bibliographystyle{unsrt}

\end{document}